\documentclass{article} 
\usepackage[final]{colm2026_conference}

\usepackage{microtype}
\usepackage{graphicx}
\usepackage{wrapfig}
\usepackage{subcaption}
\usepackage{booktabs} 

\usepackage{hyperref}
\usepackage{tikz}
\usetikzlibrary{positioning, arrows.meta, shapes.geometric, calc, decorations.pathmorphing, patterns, shadows, chains, backgrounds}

\usepackage{url}
\usepackage{lineno}

\definecolor{darkblue}{rgb}{0, 0, 0.5}
\hypersetup{colorlinks=true, citecolor=darkblue, linkcolor=darkblue, urlcolor=darkblue}

\usepackage{amsmath}
\usepackage{amssymb}
\usepackage{mathtools}
\usepackage{amsthm}
\usepackage{enumitem}

\usepackage[capitalize,noabbrev]{cleveref}

\usepackage[T1]{fontenc}
\usepackage{textcomp}

\theoremstyle{plain}
\newtheorem{theorem}{Theorem}[section]

\theoremstyle{definition}

\theoremstyle{remark}

\makeatletter
\renewcommand{\sectionautorefname}{\S\@gobble}
\renewcommand{\sectionautorefname}{\S\@gobble}
\renewcommand{\subsectionautorefname}{\S\@gobble}
\renewcommand{\sectionautorefname}{\S\@gobble}
\renewcommand{\subsectionautorefname}{\S\@gobble}
 \renewcommand{\appendixautorefname}{\S\@gobble}
\makeatother

\usepackage[textsize=tiny]{todonotes}

\usepackage{makecell}
\usepackage{multirow}

\usepackage{algorithm}
\usepackage{algorithmic}

\usepackage{listings}
\usepackage{xcolor}

\definecolor{codebg}{rgb}{0.95,0.95,0.95}
\definecolor{codegray}{rgb}{0.5,0.5,0.5}

\lstdefinelanguage{json}{
    basicstyle=\normalfont\ttfamily\small,
    numbers=none,
    numberstyle=\scriptsize,
    stepnumber=1,
    numbersep=8pt,
    showstringspaces=false,
    breaklines=true,
    frame=lines,
    backgroundcolor=\color{gray!5}, 
    string=[s]{"}{"},
    comment=[l]{//},
    morecomment=[s]{/*}{*/},
    stringstyle=\color{blue}, 
    literate=
     *{0}{{{{\color{darkgray}0}}}}{1}
      {1}{{{{\color{darkgray}1}}}}{1}
      {2}{{{{\color{darkgray}2}}}}{1}
      {3}{{{{\color{darkgray}3}}}}{1}
      {4}{{{{\color{darkgray}4}}}}{1}
      {5}{{{{\color{darkgray}5}}}}{1}
      {6}{{{{\color{darkgray}6}}}}{1}
      {7}{{{{\color{darkgray}7}}}}{1}
      {8}{{{{\color{darkgray}8}}}}{1}
      {9}{{{{\color{darkgray}9}}}}{1}
      {:}{{{{\color{black}{:}}}}}{1}
      {,}{{{{\color{black}{,}}}}}{1}
      {\{}{{{{\color{black}{\{}}}}}{1}
      {\}}{{{{\color{black}{\}}}}}}{1}
      {[}{{{{\color{black}{[}}}}}{1}
      {]}{{{{\color{black}{]}}}}}{1},
}

\usepackage{amsmath,amsfonts,bm}

\def\eqref#1{equation~\ref{#1}}

\def\1{\bm{1}}

\DeclareMathAlphabet{\mathsfit}{\encodingdefault}{\sfdefault}{m}{sl}
\SetMathAlphabet{\mathsfit}{bold}{\encodingdefault}{\sfdefault}{bx}{n}

\newcommand{\ensuretext}[1]{#1}
\newcommand{\marker}[2]{\ensuremath{^{\textsc{#1}}_{\textsc{#2}}}}
\newcommand{\arkcomment}[3]{\ensuretext{\textcolor{#3}{[#1 #2]}}}
\definecolor{bananayellow}{rgb}{1.0, 0.88, 0.21}

\newcommand{\phil}[1]{\arkcomment{\marker{Phil}{K}}{#1}{red}}

\newcommand{\joseph}[1]{\arkcomment{\marker{Jose}{A}}{#1}{orange}}

\title{Encode Once, Decode Never: Reusing Audio LM Internals for Efficient Temporal Localization}

\author{Joseph An,$^\ast$ Phillip Keung,$^\ast$ Jiaqi Wang,$^\ast$ Orevaoghene Ahia$^\ast$ and Noah A. Smith$^{\ast\dagger}$\\ $^\ast$University of Washington \quad  $^\dagger$Allen Institute for Artificial Intelligence \\
\texttt{\{anjo0,pkeung\}@uw.edu}
}

\begin{document}

\ifcolmsubmission
\linenumbers
\fi

\maketitle

\begin{abstract}
Audio language models process input audio into rich frame-level representations, but the standard approach to temporal localization generates timestamps as sequences of text tokens, which discards the frame-level representations in favor of autoregressive decoding. However, generating timestamps as tokens is slow and not parallelizable, and tends to hallucinate when producing timestamps outside the training distribution. We propose \emph{internal frame-level reuse}, a method that trains audio LMs to reuse their own internal audio representations for temporal localization directly, bypassing token generation altogether. We introduce a lightweight prediction head trained via different frame-level objectives: a binary frame classifier and a novel inhomogeneous Poisson process (IHP) loss that models temporal event intensity. Across word localization, speaker diarization, and event localization tasks, our approach can achieve a \textgreater$50\times$ inference speedup over token-based generation and demonstrates robust length generalization, maintaining high accuracy on out-of-distribution audio durations where token-based models collapse completely. We find that reusing audio frame-level representations yields localization accuracies comparable to (and often better than) finetuned token-based baselines.

\end{abstract}

\section{Introduction}
Audio language models are now widely used for various audio understanding problems, including temporal tasks such as phoneme and word alignment to audio \citep{garofolo1993timit}, speaker diarization \citep{Anguera2006_SpeakerDiarization_Review}, and audio event localization \citep{goel2025audio, ghosh2025audio, gemini2023multimodal, kimiteam2025kimiAudio, xu2025qwen3Omni, xie2025audioreasoner, wijngaard2025audsemthinker}. These tasks require reference to precise timestamps in the audio: for example, a question like ``When did the dog start barking?'' or ``When did the audience start laughing during the podcast?'' may need to be answered in units of milliseconds relative to the start of the audio recording. 
While automatic speech recognition (ASR) combined with forced alignment is a common approach for recovering word-level timestamps \citep{McAuliffe2017MontrealFA}, it is fundamentally limited to speech content and cannot generalize to the broad range of acoustic events that arise in real-world audio. 
Using an audio LM which can flexibly handle arbitrary audio understanding tasks to produce timestamps is a natural goal, but the standard approach of generating timestamps as a sequence of text tokens suffers from two fundamental problems rooted in the use of autoregressive decoding.

First, autoregressive text generation is a slow, sequential process. Audio LMs are often used to localize events such as speaker turns, overlapping speech, laughter, applause, environmental sounds, and music boundaries without requiring a separate specialized model for each type of event. A one-hour podcast of fine-grained event annotations can easily require tens of thousands of timestamp tokens, since generating a single start-end timestamp like ``[2.39, 7.11]'' uses 7--10 tokens. This replaces what should be a simple \emph{pointer} to the right audio frame with an expensive sequential \emph{decoding} process. The token-based approach fails to exploit the rich temporal information already computed and aligned internally by the audio LM when processing the audio input.

Second, autoregressive timestamp generation is brittle and prone to hallucination. Because the text decoder is unconstrained, it can generate timestamps that are syntactically correct but entirely ungrounded in the audio content \citep{ahia2025blabbrutallylongaudio, sakshi2025mmau}. This problem is especially severe for \emph{length generalization}: audio LMs tend to memorize the timestamps seen during training and fail dramatically when predicting timestamps beyond the training window (e.g., if all training audio is shorter than $n$ minutes, the model will not produce positions beyond $n$ minutes). Because the decoder does not learn to map audio features to textual representations of time beyond its training distribution, the generated timestamps for these later segments are effectively ungrounded.

The text-only approach stands in sharp contrast to the long history of models in speech recognition that use \emph{frame-level objectives} such as senone classification in DNN-HMM systems, connectionist temporal classification (CTC, \citealp{10.1145/1143844.1143891}), and RNN-Transducer models \citep{graves2012sequencetransductionrecurrentneural}. Concurrently, a separate line of research has developed LMs that are capable of tool use, enabling them to call external functions to augment their capabilities \citep{NEURIPS2023_d842425e, patil2024gorilla, 10943671, hao2023toolkengpt}. However, a major drawback of tool use is that it involves external models and API calls, which imposes latency and maintenance costs.

We propose a new approach, \emph{internal frame-level reuse}, that performs frame-level temporal localization and obviates the need for tool-augmented modeling in audio LMs. The key observation is simple: the audio LM has already encoded the input audio into a sequence of frame-level representations. Rather than forcing the model to re-derive temporal information through expensive token generation, we train a lightweight prediction head to read timestamps \emph{directly} from these existing representations. No external model is needed for temporal localization.
In our framework, the model learns to reuse its audio representations (including decoder outputs that are normally discarded, because they are not needed for token generation) to produce a probability distribution over audio frames for a given audio event query and to invoke external code to extract the relevant timestamps directly. The audio LM's decoder, conditioned on a text prompt (e.g., ``When did the second speaker start talking?''), learns to produce a frame-level alignment over its own audio representations. Simple, deterministic Python functions (e.g., peak detection, thresholding) are then applied to extract precise timestamps from this output distribution.

Our work draws inspiration from work on vision LMs like Molmo and HiMTok \citep{clark2026molmo2openweightsdata, deitke2024molmopixmoopenweights, wang2025himtoklearninghierarchicalmask}, where the model learns precise spatial and temporal localization in visual inputs (images and videos). In Molmo, for example, the model is trained to generate pixel-level x-y coordinates and temporal indices corresponding to objects or events as part of its chain-of-thought and not through the use of external tools. In addition, recent work like ChronusOmni \citep{Chen2025ChronusOmniIT} also generates times and locations without external assistance, and uses an auxiliary reinforcement learning task to improve temporal localization in video.

We explore two approaches to model the timestamps: 1) a discriminative binary loss for the frame where the audio event occurs, and 2) a novel formulation based on inhomogeneous Poisson processes (IHPs) that models the continuous-time intensity of event occurrence, providing a principled probabilistic framework for temporal localization. Our experiments span three temporal tasks: i) word localization, ii) speaker diarization, and iii) audio event localization. Our key findings are as follows:
\begin{itemize}[noitemsep, topsep=0pt]
    \item Reusing audio representations can be $\mathord{\sim}50\times$ faster than autoregressive generation because it uses a single parallel pass over audio frames instead of sequential decoding.
    \item Internal frame-level reuse substantially reduces hallucinated timestamps by grounding predictions directly in audio frames rather than generating unconstrained text tokens. In out-of-distribution scenarios where token-based models collapse to near-zero accuracy, our approach maintains over 85\% accuracy.
    \item The Poisson loss achieves accuracies and mean absolute deviations (MADs) that are comparable to and often better than token-based generation across all tasks, while providing the efficiency and robustness gains above.
    \item The binary frame-level loss is simple and effective, but the Poisson loss performs better across all tasks, highlighting the benefit of a specialized temporal loss.
\end{itemize}
In sum, our approach demonstrates that the standard practice of generating timestamps autoregressively is both slow and potentially counterproductive: directly reusing the audio LM's existing representations yields faster, robust, and more accurate temporal localization.

Our code can be found at \url{https://github.com/inkitori/taudio}. The repository also contains links and references to any unique datasets that we constructed for training and evaluation, which are hosted on HuggingFace.

\section{Internal Frame-Level Reuse}


\begin{figure*}[htbp]
    \centering
    
    \begin{subfigure}{0.48\textwidth}
        \centering
        \resizebox{\textwidth}{!}{
\begin{tikzpicture}[
    font=\sffamily,
    >=Stealth,
    node distance=1.5cm,
    block/.style={draw, rounded corners, minimum height=1cm, minimum width=2.5cm, align=center, fill=white, drop shadow},
    encoder/.style={block, fill=blue!10, draw=blue!40},
    decoder/.style={block, fill=green!10, draw=green!40, minimum width=8cm},
    head/.style={circle, draw=orange!50, fill=orange!10, inner sep=2pt, minimum size=0.8cm},
    tensor/.style={draw, fill=gray!20, minimum height=1.5cm, minimum width=0.4cm},
    label_text/.style={font=\small\bfseries, align=center},
    connection/.style={->, thick, gray!80}
]

    \node[draw, dashed, fill=yellow!5, align=center, inner sep=8pt] (prompt) at (-3, 0) {Prompt \\ \textit{"When did the dog bark?"}};

    \node[align=center] (audio_label) at (2, -0.7) {Audio Waveform ($A$)};
    \draw[blue!70, thick] (0.5,0) sin (1,0.3) cos (1.5,0) sin (2,-0.3) cos (2.5,0) 
                         sin (3,0.5) cos (3.5,0) sin (4,-0.5) cos (4.5,0);
    
    \node (waveform_center) at (2,0.2) {};

    \node[encoder, above=0.8cm of waveform_center, minimum width=6cm] (encoder) {Audio Encoder};
    \node at ([shift={(1.5cm, 1.25cm)}] encoder) {$E=\{e_1,...,e_T\}$};
    
    \draw[connection] (waveform_center) -- (encoder.south);

    
    \node[decoder, above=1.5cm of encoder] (decoder) {LLM Decoder};
    
    \draw[connection] (prompt.north) |- (decoder.west);
    
    \draw[connection] (encoder.north) --  (decoder.south) ;

    
    \coordinate (d1_pos) at ($(decoder.north) + (-3, 0.5cm)$);
    \coordinate (dt_pos) at ($(decoder.north) + (0, 0.5cm)$);
    \coordinate (dT_pos) at ($(decoder.north) + (3, 0.5cm)$);

    \node[tensor, label=left:$d_1$] (d1) at (d1_pos)[anchor=south] {};
    \node[head, above=.5cm of d1, label=left:{\scriptsize Projection + Sigmoid}] (head1) {$\sigma$};
    \node[above=0.5cm of head1] (p1) {$p_1$};
    
    \node[tensor, label=left:$d_t$] (dt) at (dt_pos)[anchor=south]{};
    \node[head, above=0.5cm of dt] (headt) {$\sigma$};
    \node[above=0.5cm of headt] (pt) {$p_t$};
    
    \node[tensor, label=left:$d_T$] (dT) at (dT_pos) [anchor=south]{};
    \node[head, above=0.5cm of dT] (headT) {$\sigma$};
    \node[above=0.5cm of headT] (pT) {$p_T$};

    \node at ($(d1)!0.5!(dt)$) {$\dots$};
    \node at ($(dt)!0.5!(dT)$) {$\dots$};

    \draw[connection] (decoder.north -| d1.south) -- (d1.south);
    \draw[connection] (d1.north) -- (head1.south);
    \draw[connection] (head1.north) -- (p1.south);

    \draw[connection] (decoder.north -| dt.south) -- (dt.south);
    \draw[connection] (dt.north) -- node[right, font=\tiny] { } (headt.south); 
    \draw[connection] (headt.north) -- (pt.south);

    \draw[connection] (decoder.north -| dT.south) -- (dT.south);
    \draw[connection] (dT.north) -- (headT.south);
    \draw[connection] (headT.north) -- (pT.south);

    \node[above=0.5cm of p1, text=red!70] (y1) {$y_1=0$};
    \node[above=0.5cm of pt, text=blue!70] (yt) {$y_t=1$};
    \node[above=0.5cm of pT, text=red!70] (yT) {$y_T=0$};

    \node[draw=red!50, fill=red!5, rounded corners, above=1.2cm of pt] (loss) {\textbf{Binary Class. Loss } $\sum_i -[y_i \log(p_i) + (1 - y_i) \log(1 - p_i)]$};

    \draw[dashed, red!40] (p1.north) -- (loss.south);
    \draw[dashed, red!40] (y1.south) -- (p1.north); 
    
    \draw[dashed, red!40] (pt.north) -- (loss.south);
    \draw[dashed, red!40] (yt.south) -- (pt.north);
    
    \draw[dashed, red!40] (pT.north) -- (loss.south);
    \draw[dashed, red!40] (yT.south) -- (pT.north);

\end{tikzpicture}
        }
        \caption{Binary frame-level prediction. $E = \{e_1,...,e_T\}$ are the audio representations from the audio encoder, and $\{d_1,...,d_T\}$ are the audio representations from the decoder's outputs. $y_t=1$ is the binary label where the event corresponding to the prompt occurred, all other labels are 0.}
        \label{fig:binary_subfig}
    \end{subfigure}
    \begin{subfigure}{0.48\textwidth}
        \centering
        \resizebox{\textwidth}{!}{
        \begin{tikzpicture}[
            font=\sffamily,
            >=Stealth,
            node distance=1.5cm,
            block/.style={draw, rounded corners, minimum height=1cm, minimum width=2.5cm, align=center, fill=white, drop shadow},
            encoder/.style={block, fill=blue!10, draw=blue!40},
            decoder/.style={block, fill=green!10, draw=green!40, minimum width=7cm},
            connection/.style={->, thick, gray!80}
        ]
            \node[draw, dashed, fill=yellow!5, align=center, inner sep=8pt] (prompt) at (-3.5, 0) {Prompt \\ \textit{"When did the dog bark?"}};
            \node[align=center] (audio_label) at (2, -0.7) {Audio Waveform ($A$)};
            \draw[blue!70, thick] (0.5,0) sin (1,0.3) cos (1.5,0) sin (2,-0.3) cos (2.5,0) sin (3,0.5) cos (3.5,0) sin (4,-0.5) cos (4.5,0);
            \node (waveform_center) at (2.5,0.2) {}; 

            \node[encoder, above=1cm of waveform_center, minimum width=5cm] (encoder) {Audio Encoder};
            \draw[connection] (2.5, 0.5) -- (encoder.south);
            \node at ([shift={(1.5cm, 1.25cm)}] encoder) {$E=\{e_1,...,e_T\}$};
            \node[decoder, above=1.5cm of encoder] (decoder) {LLM Decoder};
            \draw[connection] (prompt.north) |- ([yshift=-0.0cm]decoder.west);
            \draw[connection] (encoder.north) -- (decoder.south);

            \node[above=2cm of decoder, font=\ttfamily\large, align=center] (output_text) {\{"time": 1.21\}};
            \node[above=0.2cm of output_text] {\textbf{JSON Output}};
            \draw[connection] (decoder.north) -- (output_text.south) node[midway, right, align=left, font=\footnotesize, xshift=0.2cm] {Autoregressive\\Generation};
        \end{tikzpicture}
        
        }
        
        \caption{Direct text generation.}
        \vspace{1.4cm}
    \end{subfigure}

    \caption{Frame-level prediction versus standard autoregressive generation for predicting timestamps in audio. At inference time, we extract the predicted timestamp from the binary frame-level probabilities $p_1,..,p_T$ via argmax.}
    \label{fig:basic_diagram}
\end{figure*}
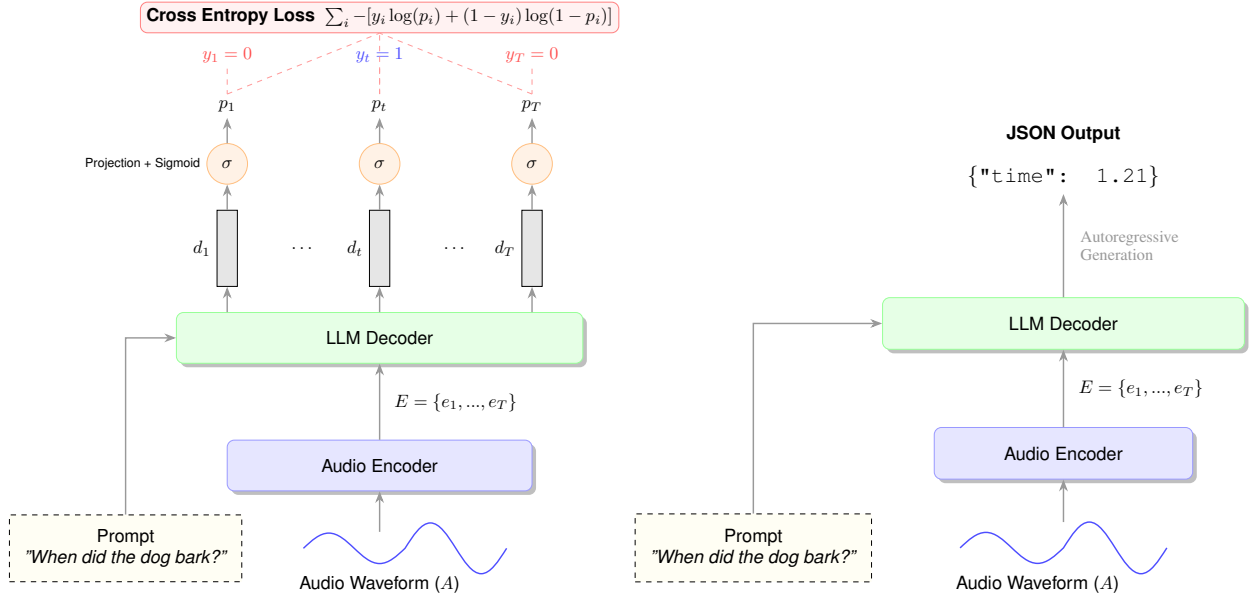

Given the flexibility language models have shown for generating so many kinds of output, and given that timestamps referencing audio input can easily be represented as text symbols, it is natural to default to direct generation of timestamps as simply ``more text'' generated autoregressively.
Our approach instead learns to construct a probability distribution aligned with the model's own internal representations of the input audio (i.e., ``frame-level reuse''). This section details the technical implementation of this mechanism.

The LM's audio representations,  $D = \{d_1, d_2, \dots, d_T\}$,  are conditioned on both the text prompt and the encoded input audio.  We attach a lightweight prediction head to $D$. Using annotated data (described in \S\ref{sec:datasets}), this head is trained to transform the decoder's output states into a probability distribution over the temporal frames, where we interpret the probability assigned to frame $t$ as the probability that the target event (referenced in the prompt) occurs at frame $t$. Simple, deterministic functions can then post-process this distribution to extract the final timestamps.

We explore two distinct frame-level loss functions, each making different assumptions about the temporal nature of the events. The first is a simple, reweighted binary cross-entropy loss, which treats each frame as an independent classification task (i.e., the frame where the event actually occurs in the training instance is a positive instance, and all the others are negative instances). The second is a novel loss function derived from the theory of inhomogeneous Poisson processes, which models the continuous-time intensity of an event occurring and provides a principled probabilistic framework for temporal localization.  We present each in turn.

\subsection{Binary Frame-Level Loss}
\label{sec:binary_loss}


In Figure \ref{fig:binary_subfig}, we illustrate binary frame-level reuse. We assume an audio LM architecture consisting of an audio encoder and a text-prompted decoder.

\begin{enumerate}
    \item The audio encoder processes a raw audio waveform $A$ into a sequence of audio feature representations, $E = \{e_1, e_2, \dots, e_T\}$, where $T$ is the total number of frames.
    \item The decoder, conditioned on a text prompt $P$ (which describes the target event), ingests the encoder audio representations $E$ to produce a sequence of decoder audio representations, $D = \{d_1, d_2, \dots, d_T\}$. \textbf{\emph{D} refers to the outputs from the final layer of the decoder}, not the intermediate representations from any other layer.
\end{enumerate}

Instead of using $D$ to generate timestamps in text form, we attach a  binary classification head (i.e., a parameter vector followed by a sigmoid activation) to each frame $d_t$. This head is trained to predict the probability $p_t$ that frame $t$ contains the target event.
The ground truth for this task, during supervised training, is a binary sequence $Y = \{y_1, y_2, \dots, y_T\}$, where $y_t = 1$ if the event described in $P$ is present at frame $t$, and $y_t = 0$ otherwise.

The usual binary cross-entropy loss for a single frame $t$ of an audio recording is:
\begin{equation}
    l_{t} = -[y_{t} \log(p_{t}) + (1 - y_{t}) \log(1 - p_{t})]
\end{equation}
where $p_{t}$ is the model's predicted probability that $y_{t} = 1$.

Of course, most frames will have $y_t = 0$.  We correct the resulting severe class imbalance  by class reweighting. For example, for a frame rate of one frame per 40 ms of audio, a 30-second audio clip would be represented by 750 frames. However, the target events would only be present in a few frames. In the case of keyword spotting, the start and end of a word would be just 2 events out of 750 frames.

We use a class weight $w$, which is set to be equal to the ratio between the negative class frequency to the positive class frequency to ensure a balanced contribution to the total loss.

During inference, the model produces, as numerical (not textual) outputs,  probabilities $\{p_{1}, p_{2}, \dots, p_{T}\}$ for each 40 ms frame of audio. To extract $k$ timestamps, we rank the probabilities for all frames in descending order and select the top-$k$ values. These indices correspond to the frames where the model has the highest confidence that the event described in the prompt $P$ has occurred.  The indices are converted into timestamps and then output textually. 

We assume that the number of events $k$ is known. For common use cases such as extracting start-of-word timestamps from a transcript, the number of words $k$ can be inferred from the generated transcript. In general, the event count prediction can be provided cheaply by the audio LM at decoding time, either as the byproduct of model inference (as in the case of audio event prediction or the transcription of speech) or simply by prompting the model to predict the total event count directly.

\subsection{Poisson Process Frame-level Loss}
\label{sec:poisson_loss}

While the binary frame-level loss assumes that events are conditionally independent across time steps, events in audio often exhibit strong dependencies over time. Furthermore, binary classification requires tuning the class reweighting due to label imbalance and struggles with ``fuzzy'' or imprecise annotations. To address these limitations, we propose a novel formulation that models event timestamps using an \emph{inhomogeneous Poisson process} (IHP). This probabilistic framework naturally handles sparsity through intensity modeling and optimizes for the likelihood of the event sequence rather than making hard, independent decisions per frame. To the best of our knowledge, this is the first application of such processes to temporal localization in audio LMs.

\paragraph{The Inhomogeneous Poisson Process}
The IHP is a stochastic counting process governing the occurrence of events over continuous time. The probability of an event occurring is defined by a time-varying intensity function, or \emph{hazard rate}, $\lambda(t) > 0$. Loosely speaking, $\lambda(t)$ is the (instantaneous) rate of events at time $t$.

We model the ground truth timestamps as the arrival times of such a process over the domain $[0, T]$, where $T$ is the audio duration in frames.

\paragraph{Parameterization and Spline Construction}
We parameterize the event intensity $\lambda(t)$ using the sequence of outputs from the audio LM's decoder, denoted $D = \{d_1, \dots, d_T\}$. We project each frame-level representation $d_k$ to a non-negative scalar rate $\lambda_k$ via a learned projection head $h$:
\begin{equation}
    \log \lambda_k = h^\top d_k
\end{equation}

\begin{wrapfigure}{R}{0.48\textwidth}
    \centering
    \resizebox{0.48\textwidth}{!}{%
    \begin{tikzpicture}[
    font=\sffamily,
    >=Stealth,
    node distance=1.5cm,
    block/.style={draw, rounded corners, minimum height=1cm, minimum width=2.5cm, align=center, fill=white, drop shadow},
    encoder/.style={block, fill=blue!10, draw=blue!40},
    decoder/.style={block, fill=green!10, draw=green!40, minimum width=8cm},
    head/.style={circle, draw=orange!50, fill=orange!10, inner sep=2pt, minimum size=0.8cm},
    tensor/.style={draw, fill=gray!20, minimum height=1.5cm, minimum width=0.4cm},
    label_text/.style={font=\small\bfseries, align=center},
    connection/.style={->, thick, gray!80},
    loss_box/.style={draw=red!50, fill=red!5, rounded corners, align=center}
]

    \node[draw, dashed, fill=yellow!5, align=center, inner sep=8pt] (prompt) at (-3, 0) {Prompt \\ \textit{"When did the dog bark?"}};

    \node[align=center] (audio_label) at (2, -0.7) {Audio Waveform ($A$)};
    
    \draw[blue!70, thick] (0.5,0) sin (1,0.3) cos (1.5,0) sin (2,-0.3) cos (2.5,0) 
                          sin (3,0.5) cos (3.5,0) sin (4,-0.5) cos (4.5,0);
    
    \node (waveform_center) at (2,0.2) {};

    \node[encoder, above=0.8cm of waveform_center, minimum width=6cm] (encoder) {Audio Encoder};
    \node at ([shift={(1.5cm, 1.25cm)}] encoder) {$E=\{e_1,...,e_T\}$};
    \draw[connection] (waveform_center) -- (encoder.south);

    \node[decoder, above=1.5cm of encoder] (decoder) {LLM Decoder};
    \draw[connection] (prompt.north) |- (decoder.west);
    \draw[connection] (encoder.north) --  (decoder.south);

    \coordinate (d1_pos) at ($(decoder.north) + (-3, 0.5cm)$);
    \coordinate (dt_pos) at ($(decoder.north) + (0, 0.5cm)$);
    \coordinate (dT_pos) at ($(decoder.north) + (3, 0.5cm)$);

    \node[tensor, label=left:$d_1$] (d1) at (d1_pos)[anchor=south] {};
    \node[head, above=.5cm of d1, label=left:{\scriptsize Projection}] (head1) { };
    \node[above=0.5cm of head1] (lam1) {$\log \lambda_1$};
    
    \node[tensor, label=left:$d_t$] (dt) at (dt_pos)[anchor=south]{};
    \node[head, above=0.5cm of dt] (headt) { };
    \node[above=0.5cm of headt] (lamt) {$\log \lambda_t$};
    
    \node[tensor, label=left:$d_T$] (dT) at (dT_pos) [anchor=south]{};
    \node[head, above=0.5cm of dT] (headT) { };
    \node[above=0.5cm of headT] (lamT) {$\log \lambda_T$};

    \node at ($(d1)!0.5!(dt)$) {$\dots$};
    \node at ($(dt)!0.5!(dT)$) {$\dots$};

    \draw[connection] (decoder.north -| d1.south) -- (d1.south);
    \draw[connection] (d1.north) -- (head1.south);
    \draw[connection] (head1.north) -- (lam1.south);

    \draw[connection] (decoder.north -| dt.south) -- (dt.south);
    \draw[connection] (dt.north) -- (headt.south);
    \draw[connection] (headt.north) -- (lamt.south);

    \draw[connection] (decoder.north -| dT.south) -- (dT.south);
    \draw[connection] (dT.north) -- (headT.south);
    \draw[connection] (headT.north) -- (lamT.south);

    
    \node[right=0.3cm of lamt, text=blue!70, align=left, font=\small] (gt_label) {Event at $t$};
    \draw[->, blue!70, dashed] (gt_label) -- (lamt);

    \node[draw=orange!60, fill=orange!5, rounded corners, inner sep=4pt, above=0.5cm of lamt, align=center] (integral) {
        \textbf{Total Intensity } 
        $\Lambda(T) = \sum_{k=1}^T \lambda_k$
    };

    \node[loss_box, above=0.5cm of integral] (loss) {
        \textbf{Poisson NLL Loss}\\
        $\mathcal{L} = -\log \lambda_t +\log\Lambda(T)$
    };

    \draw[dashed, orange!80] (lam1.north) to[out=90,in=180] (integral.west);
    \draw[dashed, orange!80] (lamt.north) -- (integral.south);
    \draw[dashed, orange!80] (lamT.north) to[out=90,in=0] (integral.east);

    \draw[->, thick, gray] (integral.north) -- (loss.south);
    
    \begin{pgfonlayer}{background}
        \draw[->, dashed, blue!70, thick] (lamt.north west) to[out=110, in=240] node[left, font=\small, text=blue!70, pos=0.7, yshift=10pt] {Select $\lambda_t$} (loss.west);
    \end{pgfonlayer}

\end{tikzpicture}

}

    \caption{Inhomogeneous Poisson process frame-level training. In the simple case where only 1 timestamp is needed, the $\log\lambda_1,...,\log\lambda_T$ are used to construct a non-parametric probability density estimator over the audio frames, normalized by the total intensity $\Lambda(T) = \sum_{k=1}^T \lambda_k$.}
    \label{fig:poisson}
    \vspace{-12mm}
\end{wrapfigure}
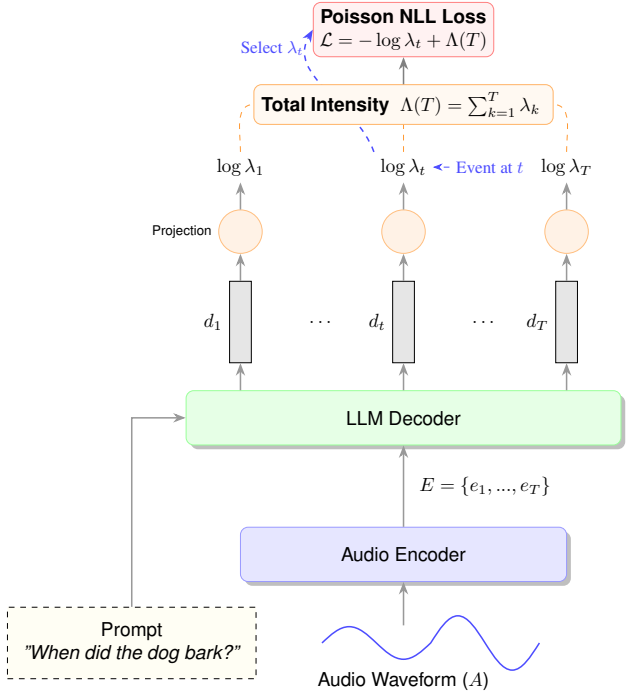
Figure \ref{fig:poisson} illustrates this parameterization in the simple case when $n=1$, i.e., where we seek to determine the timestamp for a single event of interest, such as the start of a dog's bark. The audio LM's outputs over the audio frames are converted to scalars $\log \lambda_i$ via the projection head.

We model $\lambda(t)$ as a piecewise constant spline:
\begin{equation}
    \lambda(t) = \sum^{T}_{k=1} \lambda_k \mathbb{I}\{k-1 \le t < k\}
\end{equation}
Therefore, for any time $t$ falling within the $k$th frame, the instantaneous rate is exactly equal to $\lambda_k$. 

When $n=1$, $\lambda(t)$ is like an unnormalized energy function. To convert it into a valid probability distribution, we compute the normalizing constant, which corresponds to the cumulative energy over the entire audio duration. Since the energy function is a piecewise constant spline with knots at the end of each frame, the normalizing constant is the sum of the rates: $\Lambda(T) = \int_{0}^{T}\lambda(t)dt = \sum_{i=1}^T \lambda_i$. This summation is depicted in the ``total intensity" (a.k.a, the cumulative hazard) in Figure \ref{fig:poisson}.

In general, we define the cumulative intensity at time $t$ to be $\Lambda(t) = \int_0^t \lambda(t)dt$. Note that $\Lambda(t)$ is a continuous piecewise linear spline interpolating between the knots $H_k = \sum_{j=1}^{k} \lambda_j$, since $\lambda(t)$ is itself a piecewise constant spline.

\paragraph{Loss Formulation}
The number of events is not necessarily equal to 1. Consider a training instance with $n$ event timestamps $\mathbf{t} = \{t_1, \dots, t_n\}$ such that $0 \le t_1 \le \dots \le t_n \le T$. The probability density of observing exactly these arrival times under an IHP \citep{daley2006introduction} is given by:
\begin{equation}
    p(\mathbf{t} \mid n) = \frac{n! \prod_{i=1}^n \lambda(t_i)}{\Lambda(T)^n}
\end{equation}
The negative log-likelihood loss for the sequence is therefore:
\begin{equation}
    \mathcal{L} = - \sum_{j=1}^{n} \log\lambda(t_j) + n \log\Lambda(T)
\end{equation}
In the single-event case ($n=1$) shown in Figure \ref{fig:poisson}, this simplifies to $\mathcal{L} = -\log \lambda_t + \log\Lambda(T)$, where the model essentially learns to maximize the intensity at the target frame $\lambda_t$ while suppressing the total energy $\Lambda(T)$ elsewhere.
\paragraph{Inference}
At inference time, given a predicted count of $n$ events, we wish to extract timestamps $\hat{t}_1, \dots, \hat{t}_n$. The posterior mode corresponds to the most likely specific configuration of timestamps. 

We begin with the \emph{Time Rescaling Theorem} \citep{kingman1992poisson}, which maps the complex IHP to a homogeneous Poisson process.

\begin{theorem}[Time Rescaling]
If $t_1, \dots, t_n$ is a realization of an IHP on $[0, T]$ with cumulative hazard $\Lambda(t)$, then the transformed values $z_i = \Lambda(t_i)$ form a unit-rate homogeneous Poisson process on the interval $[0, \Lambda(T)]$.
\end{theorem}

Conditioned on observing $n$ events, the sorted transformed timestamps $z_{(1)} < \dots < z_{(n)}$ are distributed as the order statistics of $n$ independent Uniform$(0, \Lambda(T))$ variables. The $i$th ordered variable $z_{(i)}$ follows a scaled Beta distribution with parameters $\alpha = i$ and $\beta = n + 1 - i$:
\begin{equation}
    p_Z(z) \propto z^{i-1} (\Lambda(T) - z)^{n-i} \quad \text{for } z \in [0, \Lambda(T)]
\end{equation}

We perform an appropriate change of variables to recover the posterior mode in the original time domain. Let $t$ denote the random variable for the $i$th timestamp in the original domain. The mapping is defined by $z = \Lambda(t)$. The probability density function $p_T(t)$ is related to $p_Z(z)$ by the Jacobian of the transformation:
\begin{equation}
    p_T(t) = p_Z(\Lambda(t)) \cdot \left| \frac{d}{dt} \Lambda(t) \right|
\end{equation}
Recalling that the derivative of the cumulative hazard is the intensity function itself, i.e., $\frac{d}{dt} \Lambda(t) = \lambda(t)$, we obtain the density for the $i$th timestamp:
\begin{equation}
    p_T(t) \propto \underbrace{\Lambda(t)^{i-1} \left(\Lambda(T) - \Lambda(t)\right)^{n-i}}_{\text{Beta contribution}} \cdot \underbrace{\lambda(t)}_{\text{Jacobian}}
\end{equation}
The posterior mode $\hat{t}_i$ is the time $t$ that maximizes this density. Taking the logarithm, we find the timestamp to be $\hat{t}_i = \operatorname*{arg\,max}_{t \in [0, T]} \log p_T(t)$. 
Since $\lambda(t)$ is a piecewise constant spline, $\Lambda(t)$ is monotonically increasing, and the beta density function is unimodal, $\log p_T(t)$ can be maximized very efficiently by evaluating it at the spline knots for each frame and at the beta mode.

In the multi-timestamp case, we found that a simpler heuristic based on an iterative mode-finding procedure and smoothing worked well, and we use this heuristic for multi-timestamp inference (see \ref{appendix:iterative}).

\section{Experimental Setup}\label{sec:datasets}

\subsection{Task and Datasets}
Our experiments focus on improving audio LM performance on human speech understanding across three tasks: (i) word localization, (ii) speaker diarization, and (iii) audio event localization. We describe the datasets used for these experiments below.

\textbf{Word Localization}: We train and test on the LibriSpeech train-clean-100, dev-clean, and test-clean datasets \citep{panayotov2015librispeech}. Ground truth timestamps are extracted using the Montreal Forced Aligner \citep{McAuliffe2017MontrealFA} via forced alignment. Each audio recording includes the transcript and the start and end timestamps for each word.

\textbf{Speaker Diarization}: We use the  Libricount dataset \citep{8506601}, a synthetic ``cocktail party'' dataset. In this setting, the task is to predict the start time of each unique speaker in a recording. Each recording contains between 1 and 10 speakers, with start times annotated for all.

\textbf{Audio Event Localization}: For identifying timestamps of audio events in the wild, we use the human sounds subset of the AudioSet dataset \citep{jort_audioset_2017}. AudioSet consists of 10-second clips from YouTube, each annotated with one or more sound categories. We focus on the strongly-labeled portion, where each acoustic event in a recording is annotated with precise start and end times. 



\subsection{Modeling}
We conducted a series of experiments to validate the effectiveness of frame-level reuse for temporal localization in audio LMs. We began by evaluating a range of open-weight audio LMs: Audio Flamingo 3 \citep{goel2025audio}, Voxtral \citep{liu2025voxtral}, and Qwen2.5-Omni \citep{xu2025qwen25omnitechnicalreport} as well as proprietary audio LMs, including Gemini 2.5 Flash \citep{gemini2023multimodal} and GPT-4o \citep{openai2024gpt4technicalreport}, in zero-shot settings across all tasks described in \autoref{sec:datasets}. 

We then finetuned the Qwen2.5-Omni 3B and 7B models on all tasks, training them to perform frame-level reuse with the binary and Poisson-based frame losses described in \autoref{sec:binary_loss} and \autoref{sec:poisson_loss}. We also finetuned these models using the usual cross-entropy loss to generate timestamp tokens under identical training conditions. (We finetuned the full model and did not use low-rank adapters. No evidence of catastrophic forgetting was found on the MMAU audio understanding benchmark; see \autoref{tab:catastrophic}.) We used compute nodes with 4 Nvidia H100 GPUs for training and inference. Hyperparameters are provided in the Appendix. 

The LM decoder first processes the prompt, followed by the audio embeddings (i.e., we place the audio embeddings after the prompt, so that the audio embeddings can be conditioned on the prompt). For Qwen2.5-Omni, the audio frames correspond to a 40ms stride and labels are attached to the audio frames in a time-synchronous fashion, and the audio hidden states are allowed to attend to the prompt. 

\paragraph{Evaluation Metrics} We evaluate the quality of generated timestamps with \textit{frame-level accuracy} (whether predicted timestamp falls within 100 ms of the ground truth) and \textit{mean absolute deviation} (MAD). For binary frame-level loss, the middle of the chosen frame is taken as the predicted timestamp.


\section{Results}

We present quantitative results evaluating frame-level reuse across multiple audio temporal localization tasks. Our experiments are designed to answer the following questions, in order of importance: (1) How much faster is frame-level reuse compared to autoregressive timestamp generation? (2) Does frame-level reuse generalize to out-of-distribution audio durations where token-based models fail? (3) How does the localization accuracy of frame-level reuse compare to token-based generation?

We also note that zero-shot temporal localization performance was poor across all the models that we tested, which necessitated model finetuning for each task. The zero-shot baseline results can be found in \autoref{tab:zeroshot_results}.


\paragraph{Frame-Level Reuse Enables Efficient Generation}\label{sec:efficiency}


The primary drawback of timestamp generation with audio LMs is the computational cost of autoregressive decoding; each timestamp is generated sequentially, and each new token requires reprocessing all previously generated tokens. This leads to linearly increasing inference time as the number of timestamps to be predicted grows. Frame-level reuse avoids autoregressive decoding altogether. Audio is processed once, in parallel, and timestamps inferred directly from frame-level predictions without additional forward passes through the model.

\begin{wrapfigure}{r}{0.5\textwidth}
    \centering
    \includegraphics[width=0.48\textwidth]{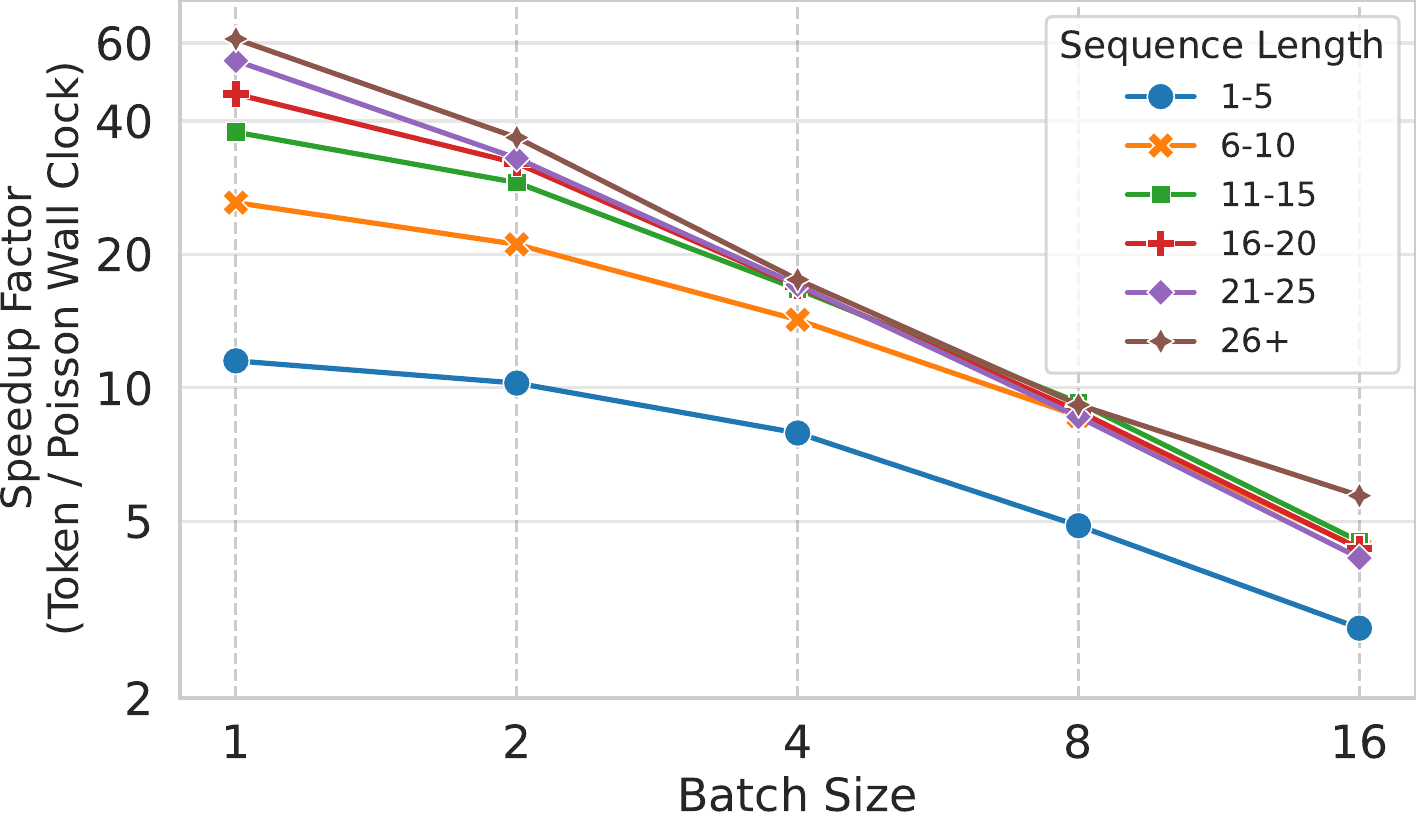}
    \caption{Speed-up factor between token-based and Poisson-based timestamp generation across batch sizes. Efficiency gains are larger when the timestamp sequences are long. Poisson frame-level timestamps are as much $60\times$ faster than token-based timestamps.}
    \label{fig:efficiency}
    \vspace{-10pt}
\end{wrapfigure}

To quantify the efficiency gains, we compare wall-clock inference time, including audio and prompt context processing, for generating multiple timestamps using token-based generation versus frame-level inference, as shown in \autoref{fig:efficiency}. We bucket examples by transcript length to analyze how runtime scales with the number of timestamps produced, as longer transcripts generally require predicting more timestamps. We observe a substantial speedup when generating timestamps using Poisson-based frame-level inference, especially when the number of words in the transcript is high. Across smaller batch sizes, inference can be $>50\times$ faster than token-based generation, while at larger batch sizes we measure at least a $\sim$3$\times$ speedup after averaging wall-clock times over the buckets.

\paragraph{Frame-Level Reuse Generalizes to Out-of-Distribution Timestamps}\label{sec:ood}
\begin{table*}[ht]
\centering

\tiny
\begin{tabular}{@{}llccccccccc@{}}
\toprule
\multirow{2}{*}{\textbf{Train}} & \multirow{2}{*}{\textbf{Test}} & \multicolumn{3}{c}{\textbf{Tokens-only}} & \multicolumn{3}{c}{\textbf{Poisson}} & \multicolumn{3}{c}{\textbf{Binary}} \\
\cmidrule(lr){3-5} \cmidrule(lr){6-8} \cmidrule(lr){9-11}
& & 20ms $\uparrow$ & 40ms $\uparrow$ & MAD $\downarrow$ & 20ms $\uparrow$ & 40ms $\uparrow$ & MAD $\downarrow$ & 20ms $\uparrow$ & 40ms $\uparrow$ & MAD $\downarrow$ \\
\midrule
\multicolumn{11}{@{}l}{\textit{Matched Distributions}} \\
\midrule
0-4s & 0-4s & 93.5\% & 97.5\% & \textbf{0.01} & \textbf{94.1\%} & \textbf{98.0\%} & \textbf{0.01} & 85.4\% & 96.3\% & 0.03 \\
0-8s & 0-8s & 93.7\% & 97.6\% & \textbf{0.01} & \textbf{93.9\%} & \textbf{98.1\%} & \textbf{0.01} & 88.2\% & 96.9\% & 0.02 \\
0-12s & 0-12s & 93.5\% & 97.8\% & \textbf{0.01} & \textbf{94.1\%} & \textbf{98.3\%} & \textbf{0.01} & 86.6\% & 95.9\% & 0.03 \\
0-16s & 0-16s & 93.7\% & 97.9\% & \textbf{0.01} & \textbf{93.9\%} & \textbf{98.4\%} & \textbf{0.01} & 85.3\% & 96.2\% & 0.04 \\
0-20s & 0-20s & 92.0\% & 96.2\% & 0.05 & \textbf{93.9\%} & \textbf{98.1\%} & \textbf{0.01} & 86.1\% & 96.3\% & 0.05 \\
\midrule
\multicolumn{11}{@{}l}{\textit{Mismatched Distributions}} \\
\midrule
0-4s & 4-8s & 0.9\% & 1.5\% & 2.35 & \textbf{92.3\%} & \textbf{97.3\%} & \textbf{0.03} & 83.5\% & 94.3\% & 0.14 \\
0-8s & 8-12s & 3.3\% & 3.6\% & 2.32 & \textbf{94.0\%} & \textbf{97.8\%} & \textbf{0.03} & 86.7\% & 96.3\% & 0.09 \\
0-12s & 12-16s & 7.4\% & 8.5\% & 2.00 & \textbf{91.2\%} & \textbf{96.6\%} & \textbf{0.04} & 81.0\% & 91.5\% & 0.20 \\
0-16s & 16-20s & 0.0\% & 0.0\% & 2.40 & \textbf{94.2\%} & \textbf{98.1\%} & \textbf{0.07} & 84.6\% & 95.5\% & 0.31 \\
0-20s & 20+s & 27.0\% & 28.6\% & 3.06 & \textbf{85.7\%} & \textbf{95.2\%} & \textbf{0.01} & 71.4\% & 87.3\% & 0.49 \\
\bottomrule
\end{tabular}
\caption{Librispeech word alignment accuracies and mean absolute deviation (MAD, in seconds) under matched (e.g., train on timestamps between 0--8 seconds, test on timestamps that are between 0--8 seconds) and mismatched (e.g., train on 0-8 seconds but test on 8--12 seconds) settings. The Poisson loss generalizes to out-of-distribution timestamps, unlike the token loss.}
\label{tab:different_times}
\end{table*}
Accurately predicting timestamps for audio events beyond the range observed during training is critical for real-world deployment. Audio LMs are prone to memorization and hallucination \citep{10888384, ahia2025blabbrutallylongaudio},
producing timestamps that reflect patterns seen in their training data rather than the actual audio content during inference. To investigate this, we finetune Qwen2.5-Omni on progressively larger timestamp ranges (i.e., from 0--4 seconds up to 0--20 seconds) using token loss, Poisson frame-level loss, or binary frame-level loss and evaluate accuracy and MAD on timestamps outside the range seen during training.

As shown in \autoref{tab:different_times}, models trained with token loss perform well on in-distribution timestamps but collapse on out-of-distribution timestamps, in some cases falling to 0\% accuracy. In contrast, our frame-level losses are far more robust: the Poisson frame-level loss consistently achieves the highest accuracy and lowest MAD on out-of-distribution timestamps, maintaining over 85\% accuracy at the 40ms threshold even when tested on timestamp ranges never seen during training. These results demonstrate a fundamental advantage of grounding predictions in audio representations rather than relying on sequential token generation: the model generalizes to timestamps beyond the training range because it is directly reading from the audio frames rather than generating text that mimics the form of timestamps it has memorized.

\paragraph{Comparing Localization Accuracy for Token Generation and Frame-level Reuse}\label{sec:finetuned}

\begin{table*}[ht]
\centering

\tiny
\begin{tabular}{@{}llrrrrrrrrr@{}}
\toprule
& & \multicolumn{3}{c}{\textbf{Librispeech}} & \multicolumn{3}{c}{\textbf{Libricount}} & \multicolumn{3}{c}{\textbf{AudioSet}}  \\
\cmidrule(lr){3-5} \cmidrule(lr){6-8} \cmidrule(lr){9-11} 
\textbf{Model} & \textbf{Method} & 20ms $\uparrow$  & 40ms $\uparrow$ & MAD $\downarrow$ & 40ms $\uparrow$ & 100ms $\uparrow$ & MAD $\downarrow$ & 40ms $\uparrow$ & 100ms $\uparrow$ & MAD $\downarrow$ \\
\midrule
\multirow{3}{*}{Qwen2.5 3B} & Tokens & 91.4\% & 96.0\% & 0.08 & 60.8\% & 73.2\% & \textbf{0.13} & 40.9\% & 55.1\% & 1.15 \\
& Poisson & \textbf{93.8\%} & \textbf{97.9\%} & \textbf{0.01} & \textbf{64.8\%} & \textbf{75.2\%} & \textbf{0.13} & \textbf{41.0\%} & \textbf{59.3\%} & \textbf{1.02} \\
& Binary & 85.8\% & 96.1\% & 0.02 & 58.4\% & 72.0\% & 0.17 & \textbf{41.0\%} & 57.7\% & 1.08 \\
\midrule
\multirow{3}{*}{Qwen2.5 7B} & Tokens & 93.2\% & 97.1\% & 0.03 & 62.1\% & 75.3\% & 0.13 & 40.1\% & 56.9\% & 1.03 \\
& Poisson & \textbf{93.3\%} & \textbf{98.1\%} & \textbf{0.01} & \textbf{65.9\%} & \textbf{76.5\%} & \textbf{0.12} & \textbf{42.3\%} & \textbf{58.5\%} & \textbf{0.99}\\
& Binary & 85.7\% & 96.5\% & 0.03 & 63.6\% & 75.5\% & 0.13 & 40.6\% & 55.8\% & 1.14 \\
\bottomrule
\end{tabular}
\caption{Finetuned accuracies (Acc, \%) and MAD (seconds) for Qwen audio LLMs with and without frame-level reuse. Highest accuracies and lowest MADs are bolded. Models tuned with Poisson loss outperform models tuned with either token or binary loss.}
\label{tab:finetuned_results}
\end{table*}

Having established the efficiency and robustness advantages of frame-level reuse, we now examine localization accuracy for single events. We observe that audio LMs finetuned with frame-level reuse perform similarly (or better than) token generation.

We finetune Qwen2.5-Omni 3B and 7B models on all tasks using frame-level reuse with both binary loss (\autoref{sec:binary_loss}) and the Poisson loss (\autoref{sec:poisson_loss}). \autoref{tab:finetuned_results} summarizes the accuracies and mean absolute deviations across tasks. (See the Appendix for examples of prompts and event types.) Since all three methods achieve high accuracies at the 100ms threshold on Librispeech, we report 20ms accuracies to highlight differences between the methods.
Across all tasks and both model sizes, the Poisson frame-level loss consistently achieves the highest accuracies and lowest MADs on Librispeech, Libricount,\footnote{We examine the pyannote baseline separately in the Appendix; it struggles with the large number of speakers in Libricount.} and AudioSet, demonstrating that the efficiency advantages of frame-level reuse do not come at the cost of localization quality. Indeed, they can come with a modest improvement. 

\paragraph{Generalizing to Multiple Timestamp Generation}
Our earlier experiments focused on predicting timestamps for single events, so we next investigate whether the Poisson frame-level loss can generalize to multiple timestamps. We finetune Qwen 2.5-Omni 7B separately on the word localization task using either token-based timestamp generation or Poisson frame-level loss, but with targets modified to include timestamps for all words in each transcript rather than a single randomly selected word.

\begin{wraptable}{R}{0.55\textwidth}
   \vspace{-4mm}
\centering

\small
\resizebox{\linewidth}{!}{
\begin{tabular}{@{}lrrrrrr@{}}
\toprule
\multirow{2}{*}{\textbf{\# of Words}} & \multicolumn{3}{c}{\textbf{Tokens-only}} & \multicolumn{3}{c}{\textbf{Poisson}} \\
\cmidrule(lr){2-4} \cmidrule(lr){5-7}
& 20ms $\uparrow$ & 40ms $\uparrow$ & MAD $\downarrow$ & 20ms $\uparrow$ & 40ms $\uparrow$ & MAD $\downarrow$ \\
\midrule
1-5   & \textbf{94.6\% }& \textbf{98.1\% }& \textbf{0.01} & 94.0\% & 97.4\% & \textbf{0.01} \\
6-10  & 94.2\% & 97.3\% & \textbf{0.01} & \textbf{94.9\%} & \textbf{97.8\%} & \textbf{0.01} \\
11-15 & 94.7\% & 97.8\% & \textbf{0.01} & \textbf{95.3\%} & \textbf{98.1\%} & \textbf{0.01} \\
16-20 & 94.3\% & \textbf{97.3\%} & \textbf{0.01} & \textbf{94.5\%} & 97.1\% & \textbf{0.01} \\
21-25 & \textbf{94.5\%}& \textbf{97.6\%} & \textbf{0.01} & 94.4\% & 97.3\% & \textbf{0.01} \\
26+   & \textbf{92.0\%} & \textbf{94.9\%} & 0.11 & 91.6\% & 94.2\% & \textbf{0.02} \\
\midrule
   & \textbf{93.2\%} & \textbf{96.2\%} & 0.06 & 93.1\% & 95.8\% & \textbf{0.02} \\
\bottomrule
\end{tabular}}
\caption{Qwen2.5-Omni 7B performance on predicting start-of-word timestamps in Librispeech, stratified by number of words in the transcript. Training with either the token or Poisson loss result in very similar performance.}
\label{tab:multi_buckets}
   \vspace{-10mm}
\end{wraptable}
We report results on the full data and across transcript-length buckets in \autoref{tab:multi_buckets} (longer transcripts contain more words, requiring more timestamp predictions). Overall, Poisson frame-level inference maintains accuracy comparable to token-based inference while providing the substantial speedups reported in \autoref{sec:efficiency}.

\paragraph{Interpolating Token and Poisson Loss}
We investigate whether combining the token-based and Poisson frame-level loss can further improve model performance. Encouraging the model to generate text tokens alongside frame-level predictions might provide additional guidance for learning precise timestamps. To test this, we implemented a training objective that adds the two losses together, weighted by a tunable coefficient $\lambda$, i.e., $L = T + \lambda P$, where $T$ is the token loss and $P$ is the Poisson frame-level loss. We ran experiments with $\lambda=0.05$ (other values such as 0.01 and 1.0 showed similar results) and report the results in \autoref{tab:joint_training}. We observe that interpolating token and Poisson losses does not consistently improve performance over using the Poisson frame-level loss alone.

\begin{table*}[ht]
\centering

\tiny
\begin{tabular}{@{}lrrrrrrrrr@{}}
\toprule
& \multicolumn{3}{c}{\textbf{\makecell[c]{Librispeech \\ (Word Align.)}}} & \multicolumn{3}{c}{\textbf{\makecell[c]{Libricount \\ (Speaker Diar.)}}} & \multicolumn{3}{c}{\textbf{\makecell[c]{Audioset \\ (Human Vocal.)}}} \\
\cmidrule(lr){2-4} \cmidrule(lr){5-7} \cmidrule(lr){8-10} 
\textbf{Model} & 20ms $\uparrow$ & 40ms $\uparrow$ & MAD $\downarrow$ & 40ms $\uparrow$ & 100ms $\uparrow$ & MAD $\downarrow$ & 40ms $\uparrow$ & 100ms $\uparrow$ & MAD $\downarrow$\\
\midrule
Audio Flamingo 3 & 0.6\% & 1.4\% & 1.82 & \textbf{2.3\%} & \textbf{6.3\%} & \textbf{1.53} & 5.8\% & 10.2\% & \textbf{1.69} \\
Voxtral 3B & 0.8\% & 1.1\% & 2.52 & 0.3\% & 1.2\% & 6.02 & 1.2\% & 2.2\% & 3.74 \\
Voxtral 24B & 0.2\% & 0.5\% & 2.12 & 0.4\% & 1.4\% & 4.91 & 2.3\% & 4.2\% & 4.03 \\
Qwen2.5 3B & 0.1\% & 0.2\% & 2.59 & 0.6\% & 1.9\% & 3.71 & 3.6\% & 4.5\% & 4.46 \\
Qwen2.5 7B & 0.7\% & 1.1\% & 2.17 & 0.4\% & 1.5\% & 5.29 & 3.9\% & 5.7\% & 3.52 \\
Gemini 2.5 Flash & \textbf{7.7}\% & \textbf{13.8}\% & \textbf{0.28} & 1.1\% & 2.3\% & 2.69 & \textbf{6.6\%} & \textbf{12.7\%} & 2.30\\
GPT-4o Audio & 0.3\% & 0.5\% & 2.42 & 0.2\% & 1.1\% & 4.60 & 1.9\% & 2.8\% & 3.80 \\
\bottomrule
\end{tabular}

\caption{
Zero-shot performance of audio LMs on temporal tasks. We report accuracies (\%) at different tolerances (20ms, 40ms, 100ms) and mean absolute deviation (MAD, seconds). Models are prompted to generate timestamps directly from audio without task-specific fine-tuning. Results show that all models struggle in this zero-shot setting, with low accuracy and high MAD across tasks. 
}
\label{tab:zeroshot_results}
\end{table*}

\begin{table*}[ht]
\centering

\tiny
\begin{tabular}{@{}lrrrrrrrrr@{}}
\toprule
 & \multicolumn{3}{c}{\textbf{Librispeech}} & \multicolumn{3}{c}{\textbf{Libricount}} & \multicolumn{3}{c}{\textbf{AudioSet}} \\
\cmidrule(lr){2-4} \cmidrule(lr){5-7} \cmidrule(lr){8-10} 
& 20ms $\uparrow$ & 40ms $\uparrow$ & MAD $\downarrow$ & 40ms $\uparrow$ & 100ms $\uparrow$ & MAD $\downarrow$ & 40ms $\uparrow$ & 100ms $\uparrow$ & MAD $\downarrow$\\
\midrule
Tokens-only & 91.4\% & 96.0\% & 0.08 & 60.8\% & 73.2\% & 0.13 & 40.9\% & 55.1\% & 1.15 \\
Poisson & 93.8\% & 97.9\% & 0.01 & 64.8\% & 75.2\% & 0.13 & 41.0\% & 59.3\% & 1.02 \\
\midrule
\multicolumn{10}{@{}l}{\emph{Interpolated loss with Poisson frame-level loss and token loss}} \\
\midrule
\quad Token Output & 92.2\% & 96.4\% & 0.06 & 62.8\% & 72.7\% & 0.15 & 41.2\% & 57.5\% & 1.01 \\
\quad Poisson Output & 94.2\% & 98.1\% & 0.01 & 63.7\% & 74.0\% & 0.15 & 41.3\% & 58.5\% & 1.04 \\
\bottomrule
\end{tabular}
\caption{Comparison of Qwen2.5-Omni 3B model performance between using token loss alone, Poisson loss alone, and an interpolation of the two losses. We ran all of our experiments with $L_\text{interpolated}=L_\text{tokens}+0.05 L_\text{Poisson}$. Interpolating the two losses shows no consistent improvement on token outputs.}
\label{tab:joint_training}
\end{table*}

\section{Conclusion}

We present \emph{internal frame-level reuse}, a method that trains audio LMs to reuse their internal audio representations for temporal localization, bypassing autoregressive token generation entirely. By predicting timestamps over audio frames rather than generating them as text tokens, our approach achieves over $50\times$ faster inference and demonstrates robust length generalization, maintaining high accuracy on out-of-distribution audio durations where token-based models collapse to near-zero performance. Importantly, these efficiency and robustness gains do not come at the cost of localization quality: across word alignment, speaker diarization, and audio event localization tasks, our Poisson-based frame-level loss achieves accuracies and MADs that are comparable or better than finetuned token-based baselines. Our results emphasize the importance of grounding temporal localization in the audio frames directly. The standard practice of generating timestamps autoregressively may be unnecessary when the audio LM has already computed frame-level representations that encode the temporal information needed for localization. Our approach can generalize beyond audio to temporal localization in video, which we leave for future work.

\section*{Acknowledgments}

This research is supported by the National Artificial Intelligence Research Resource (NAIRR) Pilot and the Anvil supercomputer supported by the National Science Foundation (award NSF-OAC 2005632) at Purdue University.


\bibliography{example_paper}

@article{garofolo1993timit,
  title={TIMIT acoustic-phonetic continuous speech corpus},
  author={Garofolo, John S and Lamel, Lori F and Fisher, William M and Pallett, David S and Dahlgren, Nancy L and Zue, Victor and Fiscus, Jonathan G},
  journal={(No Title)},
  year={1993},
  publisher={Linguistic data consortium}
}

@INPROCEEDINGS{10888384,
  author={Kuan, Chun-Yi and Lee, Hung-Yi},
  booktitle={ICASSP 2025 - 2025 IEEE International Conference on Acoustics, Speech and Signal Processing (ICASSP)}, 
  title={Can Large Audio-Language Models Truly Hear? Tackling Hallucinations with Multi-Task Assessment and Stepwise Audio Reasoning}, 
  year={2025},
  volume={},
  number={},
  pages={1-5},
  doi={10.1109/ICASSP49660.2025.10888384}}

@inproceedings{jort_audioset_2017,
    title	= {Audio Set: An ontology and human-labeled dataset for audio events},
    author	= {Jort F. Gemmeke and Daniel P. W. Ellis and Dylan Freedman and Aren Jansen and Wade Lawrence and R. Channing Moore and Manoj Plakal and Marvin Ritter},
    year	= {2017},
    booktitle	= {Proc. IEEE ICASSP 2017},
    address	= {New Orleans, LA}
}

@misc{inaguma2021alignmentknowledgedistillationonline,
      title={Alignment Knowledge Distillation for Online Streaming Attention-based Speech Recognition}, 
      author={Hirofumi Inaguma and Tatsuya Kawahara},
      year={2021},
      eprint={2103.00422},
      archivePrefix={arXiv},
      primaryClass={eess.AS},
      url={https://arxiv.org/abs/2103.00422}, 
}

@misc{wang2025himtoklearninghierarchicalmask,
      title={HiMTok: Learning Hierarchical Mask Tokens for Image Segmentation with Large Multimodal Model}, 
      author={Tao Wang and Changxu Cheng and Lingfeng Wang and Senda Chen and Wuyue Zhao},
      year={2025},
      eprint={2503.13026},
      archivePrefix={arXiv},
      primaryClass={cs.CV},
      url={https://arxiv.org/abs/2503.13026}, 
}

@inproceedings{poisson_asr,
author = {Huang, Hengguan and Wang, Hao and Mak, Brian},
title = {Recurrent poisson process unit for speech recognition},
year = {2019},
isbn = {978-1-57735-809-1},
publisher = {AAAI Press},
url = {https://doi.org/10.1609/aaai.v33i01.33016538},
doi = {10.1609/aaai.v33i01.33016538},
booktitle = {Proceedings of the Thirty-Third AAAI Conference on Artificial Intelligence and Thirty-First Innovative Applications of Artificial Intelligence Conference and Ninth AAAI Symposium on Educational Advances in Artificial Intelligence},
articleno = {802},
numpages = {8},
location = {Honolulu, Hawaii, USA},
series = {AAAI'19/IAAI'19/EAAI'19}
}

@inproceedings{Mei2017TheNHA,
  title={The Neural Hawkes Process: A Neurally Self-Modulating Multivariate Point Process},
  author={Hongyuan Mei and Jason Eisner},
  booktitle={unknown},
  year={2017},
  url={http://cs.jhu.edu/~jason/papers/mei+eisner.nips17.pdf}
}

@misc{openai2024gpt4technicalreport,
      title={GPT-4 Technical Report}, 
      author={OpenAI and Josh Achiam and Steven Adler and Sandhini Agarwal and Lama Ahmad and Ilge Akkaya and Florencia Leoni Aleman and Diogo Almeida and Janko Altenschmidt and Sam Altman and Shyamal Anadkat and Red Avila and Igor Babuschkin and Suchir Balaji and Valerie Balcom and Paul Baltescu and Haiming Bao and Mohammad Bavarian and Jeff Belgum and Irwan Bello and Jake Berdine and Gabriel Bernadett-Shapiro and Christopher Berner and Lenny Bogdonoff and Oleg Boiko and Madelaine Boyd and Anna-Luisa Brakman and Greg Brockman and Tim Brooks and Miles Brundage and Kevin Button and Trevor Cai and Rosie Campbell and Andrew Cann and Brittany Carey and Chelsea Carlson and Rory Carmichael and Brooke Chan and Che Chang and Fotis Chantzis and Derek Chen and Sully Chen and Ruby Chen and Jason Chen and Mark Chen and Ben Chess and Chester Cho and Casey Chu and Hyung Won Chung and Dave Cummings and Jeremiah Currier and Yunxing Dai and Cory Decareaux and Thomas Degry and Noah Deutsch and Damien Deville and Arka Dhar and David Dohan and Steve Dowling and Sheila Dunning and Adrien Ecoffet and Atty Eleti and Tyna Eloundou and David Farhi and Liam Fedus and Niko Felix and Simón Posada Fishman and Juston Forte and Isabella Fulford and Leo Gao and Elie Georges and Christian Gibson and Vik Goel and Tarun Gogineni and Gabriel Goh and Rapha Gontijo-Lopes and Jonathan Gordon and Morgan Grafstein and Scott Gray and Ryan Greene and Joshua Gross and Shixiang Shane Gu and Yufei Guo and Chris Hallacy and Jesse Han and Jeff Harris and Yuchen He and Mike Heaton and Johannes Heidecke and Chris Hesse and Alan Hickey and Wade Hickey and Peter Hoeschele and Brandon Houghton and Kenny Hsu and Shengli Hu and Xin Hu and Joost Huizinga and Shantanu Jain and Shawn Jain and Joanne Jang and Angela Jiang and Roger Jiang and Haozhun Jin and Denny Jin and Shino Jomoto and Billie Jonn and Heewoo Jun and Tomer Kaftan and Łukasz Kaiser and Ali Kamali and Ingmar Kanitscheider and Nitish Shirish Keskar and Tabarak Khan and Logan Kilpatrick and Jong Wook Kim and Christina Kim and Yongjik Kim and Jan Hendrik Kirchner and Jamie Kiros and Matt Knight and Daniel Kokotajlo and Łukasz Kondraciuk and Andrew Kondrich and Aris Konstantinidis and Kyle Kosic and Gretchen Krueger and Vishal Kuo and Michael Lampe and Ikai Lan and Teddy Lee and Jan Leike and Jade Leung and Daniel Levy and Chak Ming Li and Rachel Lim and Molly Lin and Stephanie Lin and Mateusz Litwin and Theresa Lopez and Ryan Lowe and Patricia Lue and Anna Makanju and Kim Malfacini and Sam Manning and Todor Markov and Yaniv Markovski and Bianca Martin and Katie Mayer and Andrew Mayne and Bob McGrew and Scott Mayer McKinney and Christine McLeavey and Paul McMillan and Jake McNeil and David Medina and Aalok Mehta and Jacob Menick and Luke Metz and Andrey Mishchenko and Pamela Mishkin and Vinnie Monaco and Evan Morikawa and Daniel Mossing and Tong Mu and Mira Murati and Oleg Murk and David Mély and Ashvin Nair and Reiichiro Nakano and Rajeev Nayak and Arvind Neelakantan and Richard Ngo and Hyeonwoo Noh and Long Ouyang and Cullen O'Keefe and Jakub Pachocki and Alex Paino and Joe Palermo and Ashley Pantuliano and Giambattista Parascandolo and Joel Parish and Emy Parparita and Alex Passos and Mikhail Pavlov and Andrew Peng and Adam Perelman and Filipe de Avila Belbute Peres and Michael Petrov and Henrique Ponde de Oliveira Pinto and Michael and Pokorny and Michelle Pokrass and Vitchyr H. Pong and Tolly Powell and Alethea Power and Boris Power and Elizabeth Proehl and Raul Puri and Alec Radford and Jack Rae and Aditya Ramesh and Cameron Raymond and Francis Real and Kendra Rimbach and Carl Ross and Bob Rotsted and Henri Roussez and Nick Ryder and Mario Saltarelli and Ted Sanders and Shibani Santurkar and Girish Sastry and Heather Schmidt and David Schnurr and John Schulman and Daniel Selsam and Kyla Sheppard and Toki Sherbakov and Jessica Shieh and Sarah Shoker and Pranav Shyam and Szymon Sidor and Eric Sigler and Maddie Simens and Jordan Sitkin and Katarina Slama and Ian Sohl and Benjamin Sokolowsky and Yang Song and Natalie Staudacher and Felipe Petroski Such and Natalie Summers and Ilya Sutskever and Jie Tang and Nikolas Tezak and Madeleine B. Thompson and Phil Tillet and Amin Tootoonchian and Elizabeth Tseng and Preston Tuggle and Nick Turley and Jerry Tworek and Juan Felipe Cerón Uribe and Andrea Vallone and Arun Vijayvergiya and Chelsea Voss and Carroll Wainwright and Justin Jay Wang and Alvin Wang and Ben Wang and Jonathan Ward and Jason Wei and CJ Weinmann and Akila Welihinda and Peter Welinder and Jiayi Weng and Lilian Weng and Matt Wiethoff and Dave Willner and Clemens Winter and Samuel Wolrich and Hannah Wong and Lauren Workman and Sherwin Wu and Jeff Wu and Michael Wu and Kai Xiao and Tao Xu and Sarah Yoo and Kevin Yu and Qiming Yuan and Wojciech Zaremba and Rowan Zellers and Chong Zhang and Marvin Zhang and Shengjia Zhao and Tianhao Zheng and Juntang Zhuang and William Zhuk and Barret Zoph},
      year={2024},
      eprint={2303.08774},
      archivePrefix={arXiv},
      primaryClass={cs.CL},
      url={https://arxiv.org/abs/2303.08774}, 
}

@misc{liu2025voxtral,
      title={Voxtral}, 
      author={Alexander H. Liu and Andy Ehrenberg and Andy Lo and Clément Denoix and Corentin Barreau and Guillaume Lample and Jean-Malo Delignon and Khyathi Raghavi Chandu and Patrick von Platen and Pavankumar Reddy Muddireddy and Sanchit Gandhi and Soham Ghosh and Srijan Mishra and Thomas Foubert and Abhinav Rastogi and Adam Yang and Albert Q. Jiang and Alexandre Sablayrolles and Amélie Héliou and Amélie Martin and Anmol Agarwal and Antoine Roux and Arthur Darcet and Arthur Mensch and Baptiste Bout and Baptiste Rozière and Baudouin De Monicault and Chris Bamford and Christian Wallenwein and Christophe Renaudin and Clémence Lanfranchi and Darius Dabert and Devendra Singh Chaplot and Devon Mizelle and Diego de las Casas and Elliot Chane-Sane and Emilien Fugier and Emma Bou Hanna and Gabrielle Berrada and Gauthier Delerce and Gauthier Guinet and Georgii Novikov and Guillaume Martin and Himanshu Jaju and Jan Ludziejewski and Jason Rute and Jean-Hadrien Chabran and Jessica Chudnovsky and Joachim Studnia and Joep Barmentlo and Jonas Amar and Josselin Somerville Roberts and Julien Denize and Karan Saxena and Karmesh Yadav and Kartik Khandelwal and Kush Jain and Lélio Renard Lavaud and Léonard Blier and Lingxiao Zhao and Louis Martin and Lucile Saulnier and Luyu Gao and Marie Pellat and Mathilde Guillaumin and Mathis Felardos and Matthieu Dinot and Maxime Darrin and Maximilian Augustin and Mickaël Seznec and Neha Gupta and Nikhil Raghuraman and Olivier Duchenne and Patricia Wang and Patryk Saffer and Paul Jacob and Paul Wambergue and Paula Kurylowicz and Philomène Chagniot and Pierre Stock and Pravesh Agrawal and Rémi Delacourt and Romain Sauvestre and Roman Soletskyi and Sagar Vaze and Sandeep Subramanian and Saurabh Garg and Shashwat Dalal and Siddharth Gandhi and Sumukh Aithal and Szymon Antoniak and Teven Le Scao and Thibault Schueller and Thibaut Lavril and Thomas Robert and Thomas Wang and Timothée Lacroix and Tom Bewley and Valeriia Nemychnikova and Victor Paltz and Virgile Richard and Wen-Ding Li and William Marshall and Xuanyu Zhang and Yihan Wan and Yunhao Tang},
      year={2025},
      eprint={2507.13264},
      archivePrefix={arXiv},
      primaryClass={cs.SD},
      url={https://arxiv.org/abs/2507.13264}, 
}

@misc{xu2025qwen25omnitechnicalreport,
      title={Qwen2.5-Omni Technical Report}, 
      author={Jin Xu and Zhifang Guo and Jinzheng He and Hangrui Hu and Ting He and Shuai Bai and Keqin Chen and Jialin Wang and Yang Fan and Kai Dang and Bin Zhang and Xiong Wang and Yunfei Chu and Junyang Lin},
      year={2025},
      eprint={2503.20215},
      archivePrefix={arXiv},
      primaryClass={cs.CL},
      url={https://arxiv.org/abs/2503.20215}, 
}

@book{kingman1992poisson,
  title={Poisson Processes},
  author={Kingman, J.F.C.},
  isbn={9780191591242},
  lccn={92025532},
  series={Oxford Studies in Probability},
  url={https://books.google.com/books?id=VEiM-OtwDHkC},
  year={1992},
  publisher={Clarendon Press}
}

@book{daley2006introduction,
  title={An Introduction to the Theory of Point Processes: Volume I: Elementary Theory and Methods},
  author={Daley, D.J. and Vere-Jones, D.},
  isbn={9780387215648},
  lccn={2002026666},
  series={Probability and Its Applications},
  url={https://books.google.com/books?id=6Sv4BwAAQBAJ},
  year={2006},
  publisher={Springer New York}
}

@ARTICLE{8506601,
  author={Stöter, Fabian-Robert and Chakrabarty, Soumitro and Edler, Bernd and Habets, Emanuël A. P.},
  journal={IEEE/ACM Transactions on Audio, Speech, and Language Processing}, 
  title={CountNet: Estimating the Number of Concurrent Speakers Using Supervised Learning}, 
  year={2019},
  volume={27},
  number={2},
  pages={268-282},
  doi={10.1109/TASLP.2018.2877892}}

@inproceedings{panayotov2015librispeech,
  title={Librispeech: an ASR corpus based on public domain audio books},
  author={Panayotov, Vassil and Chen, Guoguo and Povey, Daniel and Khudanpur, Sanjeev},
  booktitle={Acoustics, Speech and Signal Processing (ICASSP), 2015 IEEE International Conference on},
  pages={5206--5210},
  year={2015},
  organization={IEEE}
}

@inproceedings{McAuliffe2017MontrealFA,
  title={Montreal Forced Aligner: Trainable Text-Speech Alignment Using Kaldi},
  author={Michael McAuliffe and Michaela Socolof and Sarah Mihuc and Michael Wagner and Morgan Sonderegger},
  booktitle={Interspeech},
  year={2017},
  url={https://api.semanticscholar.org/CorpusID:12418404}
}

@article{Li2022LearningTA,
  title={Learning to Answer Questions in Dynamic Audio-Visual Scenarios},
  author={Guangyao Li and Yake Wei and Yapeng Tian and Chenliang Xu and Ji-rong Wen and Di Hu},
  journal={2022 IEEE/CVF Conference on Computer Vision and Pattern Recognition (CVPR)},
  year={2022},
  pages={19086-19096},
  url={https://api.semanticscholar.org/CorpusID:247763132}
}

@article{Chen2025ChronusOmniIT,
  title={ChronusOmni: Improving Time Awareness of Omni Large Language Models},
  author={Yijing Chen and Yihan Wu and Kaisi Guan and Yuchen Ren and Yuyue Wang and Ruihua Song and Liyun Ru},
  journal={ArXiv},
  year={2025},
  volume={abs/2512.09841},
  url={https://api.semanticscholar.org/CorpusID:283721592}
}

@inproceedings{Wang2025TimeR1PL,
  title={Time-R1: Post-Training Large Vision Language Model for Temporal Video Grounding},
  author={Ye Wang and Boshen Xu and Zihao Yue and Zihan Xiao and Ziheng Wang and Liang Zhang and Dingyi Yang and Wenxuan Wang and Qin Jin},
  year={2025},
  url={https://api.semanticscholar.org/CorpusID:277103988}
}

@article{Li2024GroundingGPTLanguageEM,
  title={GroundingGPT:Language Enhanced Multi-modal Grounding Model},
  author={Zhaowei Li and Qi Xu and Dong Zhang and Hang Song and Yiqing Cai and Qi Qi and Ran Zhou and Junting Pan and Zefeng Li and Van Tu Vu and Zhida Huang and Tao Wang},
  journal={ArXiv},
  year={2024},
  volume={abs/2401.06071},
  url={https://api.semanticscholar.org/CorpusID:266933314}
}

@misc{deitke2024molmopixmoopenweights,
      title={Molmo and PixMo: Open Weights and Open Data for State-of-the-Art Vision-Language Models}, 
      author={Matt Deitke and Christopher Clark and Sangho Lee and Rohun Tripathi and Yue Yang and Jae Sung Park and Mohammadreza Salehi and Niklas Muennighoff and Kyle Lo and Luca Soldaini and Jiasen Lu and Taira Anderson and Erin Bransom and Kiana Ehsani and Huong Ngo and YenSung Chen and Ajay Patel and Mark Yatskar and Chris Callison-Burch and Andrew Head and Rose Hendrix and Favyen Bastani and Eli VanderBilt and Nathan Lambert and Yvonne Chou and Arnavi Chheda and Jenna Sparks and Sam Skjonsberg and Michael Schmitz and Aaron Sarnat and Byron Bischoff and Pete Walsh and Chris Newell and Piper Wolters and Tanmay Gupta and Kuo-Hao Zeng and Jon Borchardt and Dirk Groeneveld and Crystal Nam and Sophie Lebrecht and Caitlin Wittlif and Carissa Schoenick and Oscar Michel and Ranjay Krishna and Luca Weihs and Noah A. Smith and Hannaneh Hajishirzi and Ross Girshick and Ali Farhadi and Aniruddha Kembhavi},
      year={2024},
      eprint={2409.17146},
      archivePrefix={arXiv},
      primaryClass={cs.CV},
      url={https://arxiv.org/abs/2409.17146}, 
}

@misc{clark2026molmo2openweightsdata,
      title={Molmo2: Open Weights and Data for Vision-Language Models with Video Understanding and Grounding}, 
      author={Christopher Clark and Jieyu Zhang and Zixian Ma and Jae Sung Park and Mohammadreza Salehi and Rohun Tripathi and Sangho Lee and Zhongzheng Ren and Chris Dongjoo Kim and Yinuo Yang and Vincent Shao and Yue Yang and Weikai Huang and Ziqi Gao and Taira Anderson and Jianrui Zhang and Jitesh Jain and George Stoica and Winson Han and Ali Farhadi and Ranjay Krishna},
      year={2026},
      eprint={2601.10611},
      archivePrefix={arXiv},
      primaryClass={cs.CV},
      url={https://arxiv.org/abs/2601.10611}, 
}

@INPROCEEDINGS{10943671,
  author={Wang, Chenyu and Luo, Weixin and Dong, Sixun and Xuan, Xiaohua and Li, Zhengxin and Ma, Lin and Gao, Shenghua},
  booktitle={2025 IEEE/CVF Winter Conference on Applications of Computer Vision (WACV)}, 
  title={MLLM-Tool: A Multimodal Large Language Model for Tool Agent Learning}, 
  year={2025},
  volume={},
  number={},
  pages={6678-6687},
  doi={10.1109/WACV61041.2025.00650}}

@inproceedings{
hao2023toolkengpt,
title={Toolken{GPT}: Augmenting Frozen Language Models with Massive Tools via Tool Embeddings},
author={Shibo Hao and Tianyang Liu and Zhen Wang and Zhiting Hu},
booktitle={Thirty-seventh Conference on Neural Information Processing Systems},
year={2023},
url={https://openreview.net/forum?id=BHXsb69bSx}
}

@inproceedings{
patil2024gorilla,
title={Gorilla: Large Language Model Connected with Massive {API}s},
author={Shishir G Patil and Tianjun Zhang and Xin Wang and Joseph E. Gonzalez},
booktitle={The Thirty-eighth Annual Conference on Neural Information Processing Systems},
year={2024},
url={https://openreview.net/forum?id=tBRNC6YemY}
}

@inproceedings{NEURIPS2023_d842425e,
 author = {Schick, Timo and Dwivedi-Yu, Jane and Dessi, Roberto and Raileanu, Roberta and Lomeli, Maria and Hambro, Eric and Zettlemoyer, Luke and Cancedda, Nicola and Scialom, Thomas},
 booktitle = {Advances in Neural Information Processing Systems},
 editor = {A. Oh and T. Naumann and A. Globerson and K. Saenko and M. Hardt and S. Levine},
 pages = {68539--68551},
 publisher = {Curran Associates, Inc.},
 title = {Toolformer: Language Models Can Teach Themselves to Use Tools},
 url = {https://proceedings.neurips.cc/paper_files/paper/2023/file/d842425e4bf79ba039352da0f658a906-Paper-Conference.pdf},
 volume = {36},
 year = {2023}
}

@misc{graves2012sequencetransductionrecurrentneural,
      title={Sequence Transduction with Recurrent Neural Networks}, 
      author={Alex Graves},
      year={2012},
      eprint={1211.3711},
      archivePrefix={arXiv},
      primaryClass={cs.NE},
      url={https://arxiv.org/abs/1211.3711}, 
}

@inproceedings{10.1145/1143844.1143891,
author = {Graves, Alex and Fern\'{a}ndez, Santiago and Gomez, Faustino and Schmidhuber, J\"{u}rgen},
title = {Connectionist temporal classification: labelling unsegmented sequence data with recurrent neural networks},
year = {2006},
isbn = {1595933832},
publisher = {Association for Computing Machinery},
address = {New York, NY, USA},
url = {https://doi.org/10.1145/1143844.1143891},
doi = {10.1145/1143844.1143891},
booktitle = {Proceedings of the 23rd International Conference on Machine Learning},
pages = {369–376},
numpages = {8},
location = {Pittsburgh, Pennsylvania, USA},
series = {ICML '06}
}

@inproceedings{
sakshi2025mmau,
title={{MMAU}: A Massive Multi-Task Audio Understanding and Reasoning Benchmark},
author={S Sakshi and Utkarsh Tyagi and Sonal Kumar and Ashish Seth and Ramaneswaran Selvakumar and Oriol Nieto and Ramani Duraiswami and Sreyan Ghosh and Dinesh Manocha},
booktitle={The Thirteenth International Conference on Learning Representations},
year={2025},
url={https://openreview.net/forum?id=TeVAZXr3yv}
}

@misc{ahia2025blabbrutallylongaudio,
      title={BLAB: Brutally Long Audio Bench}, 
      author={Orevaoghene Ahia and Martijn Bartelds and Kabir Ahuja and Hila Gonen and Valentin Hofmann and Siddhant Arora and Shuyue Stella Li and Vishal Puttagunta and Mofetoluwa Adeyemi and Charishma Buchireddy and Ben Walls and Noah Bennett and Shinji Watanabe and Noah A. Smith and Yulia Tsvetkov and Sachin Kumar},
      year={2025},
      eprint={2505.03054},
      archivePrefix={arXiv},
      primaryClass={cs.AI},
      url={https://arxiv.org/abs/2505.03054}, 
}

@article{Anguera2006_SpeakerDiarization_Review,
  author    = {Xavier Anguera and Simon Bozonnet and Nicholas Evans and Corinne Fredouille and Gérald Friedland and Oriol Vinyals},
  title     = {Speaker Diarization: A Review of Recent Research},
  journal   = {IEEE/ACM Transactions on Audio, Speech, and Language Processing},
  year      = {2007},
  volume    = {15},
  number    = {7},
  pages     = {1412--1421},
  note      = {Survey of classical and early diarization systems}
}

@misc{xie2025audioreasoner,
  title        = {Audio‑Reasoner: Improving Reasoning Capability in Large Audio Language Models},
  author       = {Zhifei Xie and Mingbao Lin and Zihang Liu and Pengcheng Wu and Shuicheng Yan and Chunyan Miao},
  year         = {2025},
  eprint       = {2503.02318},
  archivePrefix= {arXiv},
  primaryClass = {cs.SD},
  url          = {https://arxiv.org/abs/2503.02318}
}

@misc{wijngaard2025audsemthinker,
  title        = {AudSemThinker: Enhancing Audio‑Language Models through Reasoning over Semantics of Sound},
  author       = {Gijs Wijngaard and Elia Formisano and Michele Esposito and Michel Dumontier},
  year         = {2025},
  eprint       = {2505.14142},
  archivePrefix= {arXiv},
  primaryClass = {cs.SD},
  url          = {https://arxiv.org/abs/2505.14142}
}

@article{xu2025qwen3Omni,
  title        = {Qwen3‑Omni: A Single Multimodal Model for Text, Image, Audio, and Video},
  author       = {Xu, Jin and Guo, Zhifang and Hu, Hangrui and Chu, Yunfei and Wang, Xiong and He, Jinzheng and Wang, Yuxuan and Shi, Xian and He, Ting and Zhu, Xinfa and Lv, Yuanjun and Wang, Yongqi and Guo, Dake and Wang, He and Ma, Linhan and Zhang, Pei and Zhang, Xinyu and Hao, Hongkun and Guo, Zishan and Yang, Baosong and Zhang, Bin and Ma, Ziyang and Wei, Xipin and Bai, Shuai and Chen, Keqin and Liu, Xuejing and Wang, Peng and Yang, Mingkun and Liu, Dayiheng and Ren, Xingzhang and Zheng, Bo and Men, Rui and Zhou, Fan and Yu, Bowen and Yang, Jianxin and Yu, Le and Zhou, Jingren and Lin, Junyang},
  journal      = {arXiv preprint arXiv:2509.17765},
  year         = {2025}
}

@article{kimiteam2025kimiAudio,
  title        = {Kimi‑Audio Technical Report},
  author       = {KimiTeam and Ding, Ding and Ju, Zeqian and Leng, Yichong and Liu, Songxiang and Liu, Tong and Shang, Zeyu and Shen, Kai and Song, Wei and Tan, Xu and Tang, Heyi and Wang, Zhengtao and Wei, Chu and Xin, Yifei and Xu, Xinran and Yu, Jianwei and Zhang, Yutao and Zhou, Xinyu and Charles, Y. and Chen, Jun and Chen, Yanru and Du, Yulun and He, Weiran and Hu, Zhenxing and Lai, Guokun and Li, Qingcheng and Liu, Yangyang and Sun, Weidong and Wang, Jianzhou and Wang, Yuzhi and Wu, Yuefeng and Wu, Yuxin and Yang, Dongchao and Yang, Hao and Yang, Ying and Yang, Zhilin and Yin, Aoxiong and Yuan, Ruibin and Zhang, Yutong and Zhou, Zaida},
  journal      = {arXiv preprint arXiv:2504.18425},
  year         = {2025}
}

@article{gemini2023multimodal,
  title        = {Gemini: A Family of Highly Capable Multimodal Models},
  author       = {Gemini Team and Anil, Rohan and Alayrac, Jean‑Baptiste and Yu, Jiahui and Soricut, Radu and Schalkwyk, Johan and …},
  journal      = {arXiv preprint arXiv:2312.11805},
  year         = {2023}
}

@inproceedings{
ghosh2025audio,
title={Audio Flamingo 2: An Audio-Language Model with Long-Audio Understanding and Expert Reasoning Abilities},
author={Ghosh, Sreyan and Kong, Zhifeng and Kumar, Sonal and Sakshi, S and Kim, Jaehyeon and Ping, Wei and Valle, Rafael and Manocha, Dinesh and Catanzaro, Bryan},
booktitle={Forty-second International Conference on Machine Learning},
year={2025},
url={https://openreview.net/forum?id=xWu5qpDK6U}
}

@article{goel2025audio,
title={Audio Flamingo 3: Advancing Audio Intelligence with Fully Open Large Audio Language Models},
author={Goel, Arushi and Ghosh, Sreyan and Kim, Jaehyeon and Kumar, Sonal and Kong, Zhifeng and Lee, Sang-gil and Yang, Chao-Han Huck and Duraiswami, Ramani and Manocha, Dinesh and Valle, Rafael and Catanzaro, Bryan},
journal={arXiv preprint arXiv},
year={2025}
}
\bibliographystyle{colm2026_conference}

\appendix
\onecolumn
\section{Appendix}

\subsection{Related Work}

\paragraph{Point Processes in Machine Learning}
Point processes have been applied in various problems to model the occurrence of events in time and space across diverse settings (e.g., earthquakes, cosmology, insurance, etc.); see the treatise by \citet{daley2006introduction} for a survey. \citet{Mei2017TheNHA} incorporated point processes like the Hawkes process into recurrent neural networks to model the arrival times of events in medical, social media, and financial data. \citet{poisson_asr} incorporate a latent recurrent Poisson process model in an RNN-HMM speech recognition system to account for the occurrence time of senones. The latent process attempts to compensate for errors in the timing of the labels due to discretization and forced alignment. To the best of our knowledge, we are the first to apply point processes to temporal tasks with audio LMs.

\paragraph{Internalizing Localization Tasks in Multimodal LMs}
We have also observed a trend towards internalizing localization tasks in recent publications on multimodal models, where models generate answers to auxiliary questions directly without using a specialized external model. \citet{wang2025himtoklearninghierarchicalmask} generates image segmentation masks as tokens using a vision LM and a hierarchical mask loss. \citet{deitke2024molmopixmoopenweights} and \citet{clark2026molmo2openweightsdata} generate $<$POINT$>$ tags and x-y coordinates to refer to relevant objects as part of the chain-of-thought and the final answer. 

\paragraph{Temporal Localization in Multimodal LMs} \citet{Li2022LearningTA} introduced a variant on the usual cross attention mechanism of a CNN-LSTM encoder-decoder model to jointly align the query with the audio and visual features in videos for question answering. GroundingGPT \citep{Li2024GroundingGPTLanguageEM} is one of the earlier works that pretrained and finetuned a multimodal LM with spatial and temporal tasks in mind, but the localization was limited to producing timestamps and coordinates in token form. ChronusOmni and Time-R1 \citep{Chen2025ChronusOmniIT, Wang2025TimeR1PL} also generated timestamps in token form for moment retrieval in videos and applied reinforcement learning with interval overlap as the reward function to strengthen temporal localization. We depart from these approaches by reusing existing internal representations of the audio LM for localization instead of generating tokens directly.

\paragraph{Frame-level losses in ASR Models} RNN-T \citep{graves2012sequencetransductionrecurrentneural} and CTC \citep{10.1145/1143844.1143891} have been used as frame-level losses for speech recognition in the past. However, it is well-known that these losses do not produce good temporal alignments because they suffer from the delayed token generation problem \citep{inaguma2021alignmentknowledgedistillationonline}, where the model delays the emission of non-blank tokens until it has accumulated sufficient future acoustic context to make a confident prediction, causing the predicted timestamps to occur significantly later than the actual acoustic boundaries.


\subsection{Dataset Examples}
\label{appendix:examples}

We provide examples of prompts from the datasets and the expected model outputs from token generation. 

\subsubsection{Librispeech}
\begin{itemize}
    \item \textbf{Prompt:} When does the 1st occurrence of the word 'much' occur?
    \item \textbf{Target Output:}
\end{itemize}
\begin{lstlisting}[language=json]
```json
[
    {"word": "much", "start": 0.63}
]
```
\end{lstlisting}

\subsubsection{Libricount}
\begin{itemize}
    \item \textbf{Prompt:} When does the 3rd speaker start speaking?
    \item \textbf{Target Output:}
\end{itemize}
\begin{lstlisting}[language=json]
```json
[
    {"speaker": "3", "start": 3.744}
]
```
\end{lstlisting}

\subsubsection{AudioSet}
\begin{itemize}
    \item \textbf{Prompt:} When does the 2nd occurrence of the event 'Male singing' occur?
    \item \textbf{Target Output:}
\end{itemize}
\begin{lstlisting}[language=json]
```json
[
    {"event": "Male singing", "start": 1.821}
]
```
\end{lstlisting}

\subsubsection{Librispeech (Multi-Timestamp Inference)}
\begin{itemize}
    \item \textbf{Prompt:} 
    \begin{quote}
        Transcript: \\
        mister verloc was fully responsive now \\
        Based on the transcript, output the timestamps for every word
    \end{quote}
    \item \textbf{Target Output:}
\end{itemize}
\begin{lstlisting}[language=json]
```json
[0.21, 0.49, 0.85, 1.04, 1.27, 1.86]
```
\end{lstlisting}

\subsection{Hyperparameters}

\begin{table}[h]
\vskip 0.15in
\begin{center}
\begin{tiny}
\begin{tabular}{ll}
\toprule
\textbf{Hyperparameter} & \textbf{Value} \\
\midrule
Learning Rate & $1 \times 10^{-6}$ \\
Effective Batch Size & 8 \\
Optimizer & AdamW \\
Adam $\beta_1, \beta_2$ & 0.9, 0.999 (default) \\
Weight Decay & 0.01 (default) \\
\midrule
\textbf{Training Epochs} & \\
Qwen2.5-Omni 3B (All datasets) & 3 \\
Qwen2.5-Omni 7B (General) & 6 \\
Librispeech (All sizes, except below) & 3 \\
Librispeech 7B (Multi-timestamp task) & 6 \\
\midrule
\textbf{Checkpoint Selection Metric} & \\
Librispeech & Max 20ms Accuracy \\
Libricount / AudioSet & Max 40ms Accuracy \\
\bottomrule
\end{tabular}
\end{tiny}
\end{center}
\caption{Hyperparameter settings and training configurations.}
\label{tab:hyperparameters}
\end{table}

\subsection{Catastrophic Forgetting}

Catastrophic forgetting is always a concern when finetuning a small model on a single dataset for several epochs. We examine the performance of the Qwen2.5 7B model on the MMAU task after the model was finetuned on the Librispeech multi-timestamp prediction task with the Poisson loss; see Table~\ref{tab:catastrophic}.

\begin{table}[ht]
\centering
\tiny
\begin{tabular}{l*{4}{c}}
\toprule
Models & Sound & Music & Speech & Average \\
\midrule
Qwen2-Audio-Instruct 8B 
    & 45.9 & 53.3 & 45.9 & 52.5 \\
Qwen2.5-Omni 7B (Base Model) 
    & 64.7 & 61.2 & 64.1 & 63.3 \\
Qwen2.5-Omni 7B (Poisson-Finetuned) 
    & 63.9 & 56.9 & 61.4 & 60.7 \\
\bottomrule
\end{tabular}
\caption{MMAU results on Sound, Music, Speech, and their average on the full test set.}
\label{tab:catastrophic}
\end{table}

There is no evidence of catastrophic forgetting and the accuracy drop is modest in absolute terms on the full MMAU test set (roughly --2.6\% from 63.3\% for the base model to 60.7\% for the Poisson-finetuned model).


\subsection{Can Audio LMs Predict Timestamps Zero-Shot?}\label{sec:zero_shot}

In \autoref{tab:zeroshot_results}, we present the zero-shot accuracies and mean absolute deviations across our tasks. In this setting, models are prompted to directly generate timestamps in text form without any task-specific finetuning. Overall, zero-shot performance is poor across all benchmarks, indicating that current audio language models struggle to perform timestamp generation tasks accurately. Gemini 2.5 Flash achieves the highest overall performance; however, its average accuracy across tasks remains low. Since zero-shot performance is poor, we  evaluate the performance of finetuned models instead.

\subsection{Speaker Diarization Baseline with pyannote}
\label{appendix:pyannote}

The Libricount dataset was designed for speaker counting in a challenging setting, where 1--10 speakers might appear in any given 10-second recording (i.e., up to 10 speakers might overlap in the audio). In this paper, we used Libricount data to determine the start time of each speaker to create a speaker diarization task in a `cocktail party' setting.

pyannote, which is a widely-used speaker diarization library, would be a natural choice as a baseline system to compare against. However, the off-the-shelf Community-1 model only supports a maximum of 2 overlapping speakers.

We therefore performed a baseline evaluation on a suitable subset of our Libricount test set. pyannote was used to find the start times for 1 and 2-speaker subsets of the Libricount test set, which constitutes 400 out of the 2,000 test examples. It achieved a 40 ms accuracy of 0.41, 100 ms accuracy of 0.58, and a mean absolute deviation of 0.17 seconds. We note that this underperforms our models in Table \ref{tab:finetuned_results}, despite being an easier problem than separating the speech of up to 10 speakers.

\subsection{Learning a Linear Prediction Head with a Frozen Audio LM}

\begin{table*}[ht]
\centering
\tiny
\setlength{\tabcolsep}{4pt}
\begin{tabular}{@{}lrrrrrrrrr@{}}
\toprule
& \multicolumn{3}{c}{\textbf{Librispeech}} & \multicolumn{3}{c}{\textbf{Libricount}} & \multicolumn{3}{c}{\textbf{AudioSet}} \\
\cmidrule(lr){2-4} \cmidrule(lr){5-7} \cmidrule(lr){8-10}
\textbf{Model} & 20ms $\uparrow$ & 40ms $\uparrow$ & MAD $\downarrow$ & 40ms $\uparrow$ & 100ms $\uparrow$ & MAD $\downarrow$ & 40ms $\uparrow$ & 100ms $\uparrow$ & MAD $\downarrow$ \\
\midrule
3B & 0.08 & 0.11 & 2.33 & 0.18 & 0.23 & 1.55 & 0.05 & 0.11 & 2.97 \\
7B & 0.04 & 0.06 & 2.21 & 0.14 & 0.19 & 1.79 & 0.10 & 0.19 & 2.74 \\
\bottomrule
\end{tabular}
\caption{Performance of the 3B and 7B models on Librispeech, Libricount, and AudioSet at different temporal resolutions when the audio LM is frozen and only the linear prediction head is trained.}
\label{tab:model_size_results}
\end{table*}

In this experiment, the audio LM was kept frozen, and only the linear prediction head was learned, and we used the frame-level binary classification loss. 

The audio LM acoustic features were optimized for tasks besides temporal localization (most audio tasks used to train audio LMs involve speech recognition). This underscores the need to incorporate temporal localization tasks into model training pipelines, and motivates the finetuning of the entire audio LM in our experiments.

\subsection{Iterative Mode Finding and Smoothing for Multi-timestamp Inference}
\label{appendix:iterative}
We convolve the frame-level intensities $\lambda_1, \dots, \lambda_T$ with a boxcar kernel of width $M = \lfloor 2\tau/\delta \rfloor$ frames, where $\tau$ is the evaluation tolerance and $\delta$ is the frame duration, so that the smoothed value at frame $t$ equals the total intensity within $\pm\tau$ of $t$. We then repeat $n$ times: select the frame with maximal smoothed intensity, record it as a prediction, and zero out the intensities in the surrounding window before re-smoothing, so that no event's intensity is counted twice. The $n$ selected frames, sorted, form the final timestamp predictions.

\begin{figure}[h]
    \centering
    \includegraphics[width=0.85\linewidth]{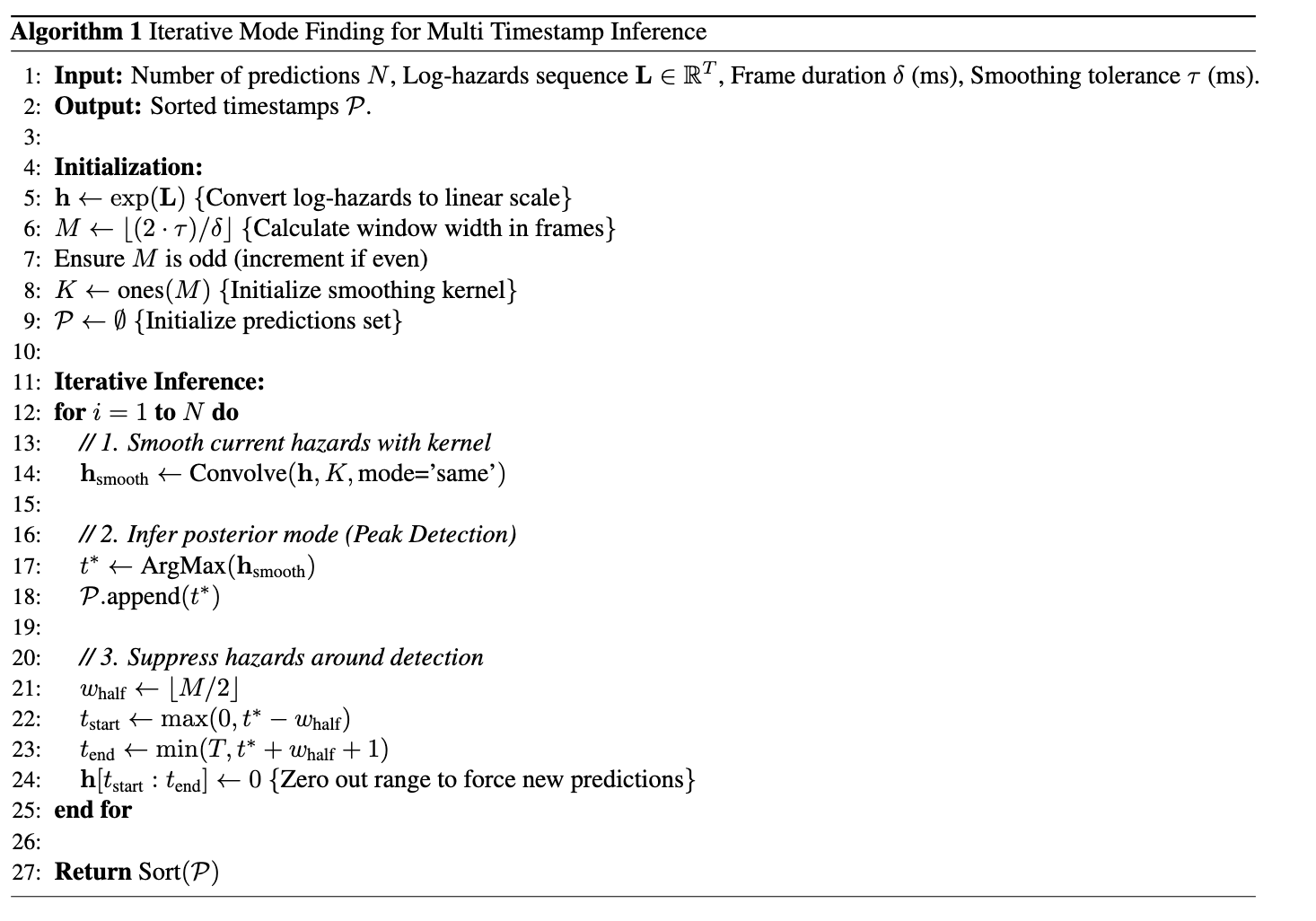}
    \caption{Multi-timestamp decoding with our iterative mode finding and smoothing heuristic.}
    \label{fig:placeholder}
\end{figure}

\end{document}